%% file: main.tex
\documentclass[sigconf]{acmart}
\usepackage{url}
\usepackage{amsmath}
\usepackage{bm}
\usepackage{algorithm}
\usepackage{algorithmicx}
\usepackage{algpseudocode}
\usepackage{mathrsfs}
\usepackage{multirow}
\usepackage{subfigure}
\usepackage{caption}
\usepackage{enumitem}
\usepackage{xcolor}

\newcommand{\algorithmicinput}{\textbf{Input:}}
\newcommand{\INPUT}{\item[\algorithmicinput]}
\newcommand{\OUTPUT}{\item[\algorithmicensure]}
\renewcommand{\algorithmicensure}{\textbf{Output:}}

\input{math_commands.tex}

\setcopyright{acmlicensed}
\copyrightyear{2018}
\acmYear{2018}
\acmDOI{XXXXXXX.XXXXXXX}

\acmConference[Conference acronym 'XX]{Make sure to enter the correct
  conference title from your rights confirmation email}{June 03--05,
  2018}{Woodstock, NY}
\acmISBN{978-1-4503-XXXX-X/2018/06}

\begin{document}

%%
%% The "title" command has an optional parameter,
%% allowing the author to define a "short title" to be used in page headers.
%\title{Deep Manifold Reshaping for Learning Tabular Data on the Web}
\title{Calibrating Tabular Anomaly Detection via Optimal Transport}
%
% The "author" command and its associated commands are used to define
% the authors and their affiliations.
% Of note is the shared affiliation of the first two authors, and the
% "authornote" and "authornotemark" commands
% used to denote shared contribution to the research.
% \author{Ben Trovato}
% \authornote{Both authors contributed equally to this research.}
% \email{trovato@corporation.com}
% \orcid{1234-5678-9012}
% \author{G.K.M. Tobin}
% \authornotemark[1]
% \email{webmaster@marysville-ohio.com}
% \affiliation{%
%   \institution{Institute for Clarity in Documentation}
%   \streetaddress{P.O. Box 1212}
%   \city{Dublin}
%   \state{Ohio}
%   \country{USA}
%   \postcode{43017-6221}
% }
\author{Hangting Ye}
% \authornotemark[1]
\affiliation{%
  \institution{School of Artificial Intelligence, \\Jilin University}
  \city{Changchun}
  \country{China}
  % \postcode{130012}
}
\email{yeht2118@mails.jlu.edu.cn}

\author{He Zhao}
% \authornotemark[1]
\affiliation{%
  \institution{CSIRO's Data61}
  \city{Sydney}
  \country{Australia}
  % \postcode{130012}
}
\email{he.zhao@data61.csiro.au}

\author{Wei Fan}
% \authornotemark[1]
\affiliation{%
  % \institution{Nuffield Department of Women's \\\& Reproductive Health, \\University of Oxford}
  \institution{University of Auckland}
  \city{Auckland}
  \country{New Zealand}
  % \postcode{130012}
}
\email{wei.fan@auckland.ac.nz}

\author{Xiaozhuang Song}
% \authornotemark[1]
\affiliation{%
  \institution{Chinese University of Hong Kong, Shenzhen}
  \city{Shenzhen}
  \country{China}
  % \postcode{130012}
}
\email{xiaozhuangsong1@link.cuhk.edu.cn}

\author{Dandan Guo}
% \authornotemark[1]
\affiliation{%
  \institution{School of Artificial Intelligence, \\Jilin University}
  \city{Changchun}
  \country{China}
  % \postcode{130012}
}
\email{guodandan@jlu.edu.cn}

\author{Yi Chang}
% \authornotemark[1]
\affiliation{%
  \institution{School of Artificial Intelligence, \\Jilin University}
  \city{Changchun}
  \country{China}
  % \postcode{130012}
}
\email{yichang@jlu.edu.cn}

\author{Hongyuan Zha}
% \authornotemark[1]
\affiliation{%
  \institution{Chinese University of Hong Kong, Shenzhen}
  \city{Shenzhen}
  \country{China}
  % \postcode{130012}
}
\email{zhahy@cuhk.edu.cn}

%%
%% By default, the full list of authors will be used in the page
%% headers. Often, this list is too long, and will overlap
%% other information printed in the page headers. This command allows
%% the author to define a more concise list
%% of authors' names for this purpose.
% \renewcommand{\shortauthors}{Trovato and Tobin, et al.}

%%
%% The abstract is a short summary of the work to be presented in the
%% article.
\begin{abstract}
  \input{0_abstract_improved}

\end{abstract}

%
% The code below is generated by the tool at http://dl.acm.org/ccs.cfm.
% Please copy and paste the code instead of the example below.
%


%
% Keywords. The author(s) should pick words that accurately describe
% the work being presented. Separate the keywords with commas.
\keywords{Do, Not, Us, This, Code, Put, the, Correct, Terms, for,
  Your, Paper}

% \received{20 February 2007}
% \received[revised]{12 March 2009}
% \received[accepted]{5 June 2009}

%%
%% This command processes the author and affiliation and title
%% information and builds the first part of the formatted document.
\maketitle

\input{1_introduction_improved}
\input{2_related_work}

\input{3_preliminary}
\input{4_method_final}
\input{5_experiments}
\input{6_conclusion}

\bibliographystyle{ACM-Reference-Format}
\bibliography{ref}

%%
%% If your work has an appendix, this is the place to put it.

\input{Appendix_final}

\end{document}

%% file: math_commands.tex
\usepackage{amsmath,amsfonts,bm}

\def\eqref#1{equation~\ref{#1}}
\def\1{\bm{1}}

\def\mP{{\bm{P}}}
\def\mQ{{\bm{Q}}}

\DeclareMathAlphabet{\mathsfit}{\encodingdefault}{\sfdefault}{m}{sl}
\SetMathAlphabet{\mathsfit}{bold}{\encodingdefault}{\sfdefault}{bx}{n}

%% file: 0_abstract_improved.tex
Tabular anomaly detection (TAD) remains challenging due to the heterogeneity of tabular data: features lack natural relationships, vary widely in distribution and scale, and exhibit diverse types. Consequently, each TAD method makes implicit assumptions about anomaly patterns that work well on some datasets but fail on others, and no method consistently outperforms across diverse scenarios. We present CTAD (Calibrating Tabular Anomaly Detection), a model-agnostic post-processing framework that enhances any existing TAD detector through sample-specific calibration. Our approach characterizes normal data via two complementary distributions—an empirical distribution from random sampling and a structural distribution from K-means centroids—and measures how adding a test sample disrupts their compatibility using Optimal Transport (OT) distance. Normal samples maintain low disruption while anomalies cause high disruption, providing a calibration signal to amplify detection. We prove that OT distance has a lower bound proportional to the test sample's distance from centroids, and establish that anomalies systematically receive higher calibration scores than normals in expectation, explaining why the method generalizes across datasets. Extensive experiments on 34 diverse tabular datasets with 7 representative detectors spanning all major TAD categories (density estimation, classification, reconstruction, and isolation-based methods) demonstrate that CTAD consistently improves performance with statistical significance. Remarkably, CTAD enhances even state-of-the-art deep learning methods and shows robust performance across diverse hyperparameter settings, requiring no additional tuning for practical deployment.

%% file: 1_introduction_improved.tex
\section{Introduction} \label{sec:intro}

Anomaly detection (AD), also known as outlier detection, is the task of identifying data points or behaviors that significantly deviate from the majority~\cite{zha2020meta}. AD is considered a critical problem in machine learning, with applications spanning various domains, including web and cyber security (e.g., intrusion detection)~\cite{ahmad2021network, cui2021security}, finance (e.g., financial fraud detection)~\citep{al2021financial, dixon2020machine}, and healthcare (e.g., rare disease diagnosis)~\cite{chandola2009anomaly, fernando2021deep}.
% Among them, the majority are designed for tabular data (i.e., no time dependency and graph structure)~\cite{pang2021deep, han2022adbench}.
In this paper, we focus on anomaly detection for tabular
data (i.e., no time dependency and graph structure), which is a very challenging problem and has been the
focus of most related works in the literature~\cite{pang2021deep, han2022adbench, pang2018learning, wang2019unsupervised, nguyen2018scalable, zong2018deep}.
In tabular anomaly detection (TAD), anomalies usually need to be manually annotated by domain experts, which is both expensive and time-consuming, and accurately marking all types of abnormal samples is usually unaffordable in practice \cite{chandola2009anomaly}.
Therefore, TAD is often implemented in one-class classification setting, i.e., only normal samples are available during training~\cite{ruff2018deep, shenkar2022anomaly, yin2024mcm}.

\noindent\textbf{The fundamental challenge.} Despite extensive research~\cite{pang2021deep, han2022adbench}, no single TAD method consistently outperforms all others across diverse datasets~\cite{han2022adbench, wolpert1997no}.
Each method makes implicit assumptions about what constitutes an anomaly~\cite{ahmed2016survey}.
These assumptions work well when the data conforms to them, but fail when violated—a common occurrence with heterogeneous tabular data.
Unlike images, text, or graphs, tabular data lacks inherent structure: features have no natural spatial or sequential relationships, and they exhibit extreme heterogeneity in distributions, scales, and types (continuous vs. categorical)~\cite{shenkar2022anomaly, ye2023uadb, ye2024ptarl}.
A method that succeeds on one tabular dataset often fails on another due to drastically different data characteristics.
This dataset-specificity problem is well-documented: \textit{there exists no universal winner for TAD}~\cite{wolpert1997no, han2022adbench}.

\noindent\textbf{Our insight.} Rather than seeking yet another assumption-based detector that will inevitably fail on some datasets, we take a fundamentally different approach: \textit{calibrate existing detectors to improve their performance across diverse data}.
The key observation is that any well-trained detector—even an imperfect one—captures some signal about normality. Our goal is to amplify this signal through adaptive, sample-specific error correction.

We introduce CTAD (Calibrating Tabular Anomaly Detection), a model-agnostic framework that enhances any TAD model by adding a calibration term $\Delta$ to its anomaly scores.
The core idea is elegant: we characterize the normal distribution through two complementary views—an empirical distribution $\mP$ via random sampling and a structural distribution $\mQ$ via K-means centroids.
For a test sample $\mathbf{x}_{test}$, we embed it into $\mP$ and measure the disruption to the compatibility between $\mP$ and $\mQ$ using Optimal Transport (OT) distance~\cite{peyre2019computational}.
Normal samples preserve compatibility (low OT), while anomalies disrupt it (high OT), providing the calibration signal $\Delta_{test}$.

\noindent\textbf{Theoretical guarantees.} We establish rigorous theoretical foundations for CTAD:
\textit{Lower bound}: The OT distance is bounded below by the test sample's distance to the nearest centroid, ensuring anomalies (far from centroids) receive higher scores than normals (near centroids).
\textit{Expected separation}: Under natural separability assumptions, we prove that anomalies systematically receive higher calibration terms than normal samples in expectation, guaranteeing improved detection.
\textit{Tight characterization}: We provide both lower and upper bounds on the OT distance, fully characterizing how the calibration term depends on the test sample's position relative to normal structure.
These results explain \textit{why} CTAD works, not just \textit{that} it works.
% \begin{itemize}[leftmargin=*, noitemsep, topsep=3pt]
% \item \textit{Lower bound}: The OT distance is bounded below by the test sample's distance to the nearest centroid, ensuring anomalies (far from centroids) receive higher scores than normals (near centroids).
% \item \textit{Expected separation}: Under natural separability assumptions, we prove that anomalies systematically receive higher calibration terms than normal samples in expectation, guaranteeing improved detection.
% \item \textit{Tight characterization}: We provide both lower and upper bounds on the OT distance, fully characterizing how the calibration term depends on the test sample's position relative to normal structure.
% \end{itemize}
% These results explain \textit{why} CTAD works, not just \textit{that} it works.

% \noindent\textbf{Empirical validation.} We conduct extensive experiments on 34 diverse tabular datasets with 5 representative TAD baselines spanning classical methods and deep learning approaches.
% CTAD significantly improves performance across all baselines, with average gains of \textcolor{blue}{x-x\% in AUC-PR and x-x\% in AUC-ROC}.
% Critically, CTAD adds negligible computational overhead (\textcolor{blue}{$<$1\%} of base detector cost) and requires no retraining—it operates purely as post-processing.

In brief, this paper makes the following contributions:

\begin{itemize}[leftmargin=*]
    \item 
    \textbf{A new paradigm for TAD}: We shift focus from designing yet another assumption-based detector to calibrating existing detectors, enabling them to adapt to diverse datasets.
    
    \item 
    \textbf{The CTAD framework}: We introduce a model-agnostic calibration approach based on optimal transport between two complementary characterizations of normality, with provable theoretical guarantees.
    
    \item 
    \textbf{Rigorous theoretical analysis}: We prove that CTAD systematically assigns higher calibration scores to anomalies than to normal samples, providing lower and upper bounds that fully characterize the method's behavior.
    
    \item 
    \textbf{Comprehensive empirical study}: We validate CTAD on 34 datasets with 7 baselines, demonstrating consistent improvements and providing detailed analysis of when and why CTAD succeeds.
\end{itemize}

%% file: 2_related_work.tex
\section{Related Work}
\label{sec:related_work}
Tabular anomaly detection approaches can be categorized into three main types: density estimation-based, classification-based, reconstruction-based and isolation-based methods.

\noindent\textbf{Density Estimation-based Methods.}  
These approaches typically estimate the normal distribution directly and measure the likelihood of a sample under the estimated distribution. Traditional methods include KDE~\cite{parzen1962estimation} and GMM~\cite{roberts1994probabilistic}. 
Additionally, some methods focus on local density estimation to detect outliers, such as the Local Outlier Factor (LOF)~\cite{breunig2000lof}. LOF quantifies the local deviation in density of a given sample compared to its neighbors, effectively identifying regions with sparse data points. During inference, samples residing in low-probability regions of the estimated distribution are flagged as anomalies.
Recently, ECOD~\cite{li2022ecod} has employed empirical cumulative distribution for this purpose.

\noindent\textbf{Classification-based Methods.}
These approaches involve discriminative models that learns a decision boundary, and assume that normal and anomaly samples are divided into different regions.
For instance, kernel-based methods~\cite{scholkopf1999support} (e.g., OCSVM) define the support of normal samples in a Hilbert space, flagging as anomalies any samples that fall outside the estimated support. Building on this idea, recent advancements have replaced kernels with deep neural networks to enhance modeling capacity~\cite{ruff2018deep}. Similarly, DROCC~\cite{goyal2020drocc} introduces a novel strategy by generating synthetic anomalous samples during training, enabling the model to learn a robust classifier on top of the one-class representation.

% This category focuses on learning a decision boundary using only normal data.
% For example, OCSVM~\citep{scholkopf1999support} creates this boundary by maximizing the margin between the input data and the origin. 
% Similarly, ~\citep{tax2004support} propose a method to learn the smallest hypersphere that encapsulates most samples, labeling any data outside this hypersphere as anomalies.

\noindent\textbf{Isolation-based Methods}. This category of methods like IForest assume anomalies are easily isolated~\cite{liu2008isolation, cao2025anomaly}. Their effectiveness diminishes when anomalies form small clusters or are adjacent to normal ones. In these cases, the isolation paths of anomalies are no longer much shorter than those of normal data, compromising the core principle of isolation. The inherent brittleness of these varied assumptions motivates our work to directly enhance detectors against the violation of their own core logic. For a comprehensive overview, we refer readers to several surveys~\cite{pang2021deep, ruff2021unifying, chandola2009anomaly, cao2025anomaly}.

\noindent\textbf{Reconstruction-based Methods.}
Other approaches focus on learning to reconstruct samples from the normal distribution, where a model’s inability to accurately reconstruct a sample serves as a proxy for detecting anomalies. A high reconstruction error indicates that the sample likely deviates from the estimated normal distribution. These methods include techniques such as Principal Component Analysis (PCA)~\cite{shyu2003novel}, various types of autoencoders~\cite{principi2017acoustic, kim2019rapp}, and Generative Adversarial Networks (GANs)~\cite{schlegl2017unsupervised}. 
More recently, self-supervised learning techniques have been explored for anomaly detection. For instance, in GOAD~\cite{bergman2020classification}, several affine transformations are applied to each sample, and a classifier is trained to identify the transformation applied. The classifier is trained exclusively on normal samples, making it likely to fail in identifying transformations applied to anomalies, as this task is assumed to be class-dependent. 
% ICL~\cite{shenkar2022anomaly} introduces a contrastive learning framework that detects anomalies by analyzing inter-feature relationships. 
MCM~\cite{yin2024mcm} proposes a learnable masking strategy, where the model reconstructs masked input values based on unmasked entries, leveraging this mechanism for anomaly detection.
Instead of relying purely on reconstruction in observation space, DRL~\cite{ye2025drl} enforces a structured latent space where each normal representation is expressed as a linear combination of fixed, orthogonal basis vectors.

%% file: 3_preliminary.tex
\section{Preliminaries}

\subsection{Problem Statement and Notations} \label{sec:problem_statement}
Following previous works~\cite{yin2024mcm, ye2025drl}, we implement tabular anomaly detection (TAD) in a one-class classification setting, i.e. only normal samples are available during training. We denote the training set as $\mathcal{D}_{train} = \{\mathbf{x}_i\}_{i=1}^N$, where $\mathbf{x}_i \in \mathcal{X} \subseteq \mathbb{R}^D$, $N$ is the number of normal samples in the training set and $D$ is the number of input features. The test set $\mathcal{D}_{test}$ includes both anomalous and normal samples. The goal of TAD is to learn a detection model $f(\mathbf{x})\to s\in\mathbb{R}$ from $\mathcal{D}_{train}$.
During inference, the detection model takes sample $\mathbf{x}_{test}\in \mathcal{D}_{test}$ as input and outputs the predicted anomaly score $s_{test}$, where a higher score indicates a higher confidence that $\mathbf{x}_{test}$ is an anomaly.
In this paper, we consider the problem of calibrating any well trained detection model's output $s_{test}$ of $\mathbf{x}_{test}\in \mathcal{D}_{test}$ by $s_{test}^* \gets s_{test} + \Delta_{test}$ to achieve better performance, where $\Delta_{test}$ is the error correction item and $s_{test}^*$ is the calibrated anomaly score.

\subsection{Optimal Transport}
\label{sec:ot}
%Optimal transport provides meaningful and sound measurements to compare data distributions~\cite{} 
Optimal transport has provided sound and meaningful measurements to compare difference of data distributions~\cite{peyre2019computational}, which has been used to solve many problems, such as computer vision~\cite{zhang2020deepemd}, text modeling ~\cite{yurochkin2019hierarchical}, adversarial robustness~\cite{sanjabi2018convergence}, and other machine learning applications~\cite{vayer2018optimal}.
We now introduce optimal transport between two discrete probability distributions.
Let us consider $\mP$ and $\mQ$ as two discrete probability distributions over an arbitrary space $\mathbb{S} \in \mathbb{R}^D$, which can be expressed as $\mP=\sum_{i=1}^{n} \mathbf{a}_i \delta_{\mathbf{x}_i}$ and $\mQ=\sum_{j=1}^{m} \mathbf{b}_j \delta_{\mathbf{y}_j}$. 
In this case, $\mathbf{a} \in \sum^{n}$ and $\mathbf{b} \in \sum^{m}$, where $\sum^{n}$ represents the probability simplex in $\mathbb{R}^n$. The OT distance between $\mP$ and $\mQ$ is defined as:
\begin{equation}
\label{ot_def}
\begin{aligned}
    \mathbf{OT}(\mP, \mQ) = \min_{\textbf{T}\in \Pi (\mP, \mQ)} \langle \textbf{T}, \textbf{C} \rangle ,
\end{aligned}
\end{equation}
where $\langle \cdot, \cdot \rangle$  is the Frobenius dot-product and $\mathbf{C} \in \mathbb{R}^{n\times m}_{\geq 0}$ is the transport cost matrix constructed by $\mathbf{C}_{ij} = \text{Distance}(\mathbf{x}_i, \mathbf{y}_j)$.
The transport probability matrix $\textbf{T} \in \mathbb{R}^{n\times m}_{\geq 0}$, which satisfies $\Pi(\mP,\mQ):= \{\textbf{T} | \sum_{i=1}^{n} \mathbf{T}_{ij}=\mathbf{b}_j,  \sum_{j=1}^{m} \mathbf{T}_{ij}=\mathbf{a}_i\}$, is learned by minimizing $\langle \textbf{T}, \textbf{C} \rangle$. 

%% file: 4_method_final.tex
\section{Methodology}

We now present \textbf{CTAD} (Calibrating Tabular Anomaly Detection), a model-agnostic post-processing framework that enhances any base anomaly detector through sample-specific calibration.
The core principle is elegant: rather than designing yet another detector with fixed assumptions about anomaly patterns, we augment existing detectors with a \emph{calibration term} that measures how much a test sample disrupts the internal consistency of normal data.
Critically, CTAD operates purely as post-processing---it requires no model retraining, no anomaly labels during deployment, and no modifications to the base detector's architecture.

\begin{figure}[!t]
\centering
\includegraphics[width=\linewidth]{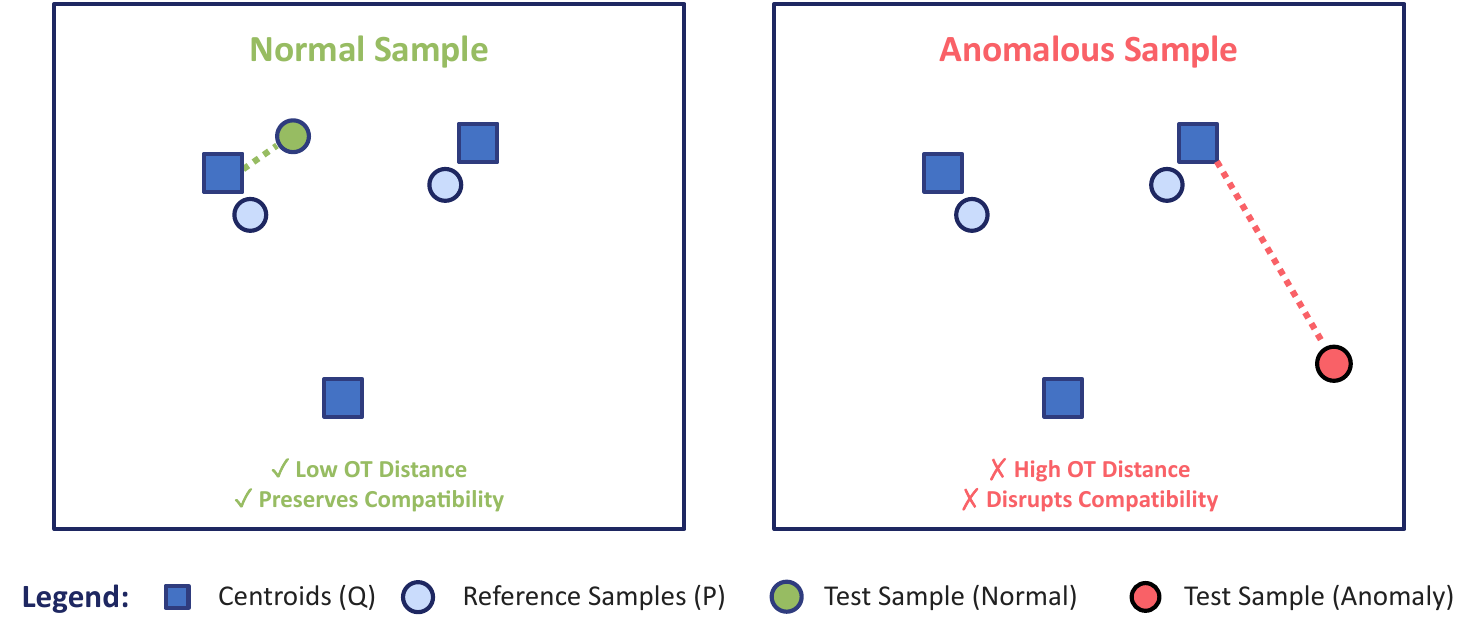}
\captionsetup{font=small}
% \captionsetup{skip=0pt}
\caption{The two-distribution philosophy of CTAD. 
We construct an empirical distribution $\mP$ by augmenting random reference samples from $\mathcal{D}_{train}$ with $\mathbf{x}_{test}$, and a structural distribution $\mQ$ from K-means centroids of $\mathcal{D}_{train}$.
\textbf{Left:} A normal test sample lies close to centroids, resulting in low OT distance and preserved compatibility.
\textbf{Right:} An anomalous test sample is far from all centroids, resulting in high OT distance and disrupted compatibility.
The dashed line illustrates the transport cost that $\mathbf{x}_{test}$ must pay to align with $\mQ$.} 
\label{fig:framework}
\vspace{-1em}
\end{figure}

\textbf{The two-distribution philosophy.}
At its heart, CTAD leverages two complementary characterizations of normality.
The first is an \emph{empirical distribution}~$\mP$, constructed by randomly sampling points from the training set and augmenting them with the test sample.
This provides a fine-grained snapshot capturing local variability in the data.
The second is a \emph{structural distribution}~$\mQ$, obtained via K-means clustering on the training set, which distills the coarse global structure into a set of representative centroids.
The key insight is geometric: for a normal test point, embedding it into~$\mP$ should preserve compatibility with~$\mQ$; for an anomaly, this compatibility deteriorates (Fig.~\ref{fig:framework} explains this intuition).
We formalize this notion of compatibility using \emph{Optimal Transport} (OT), which measures the minimal cost to align two distributions.
The OT distance between~$\mP$ and~$\mQ$ serves as our calibration signal, quantifying the structural implausibility of the test point.

\textbf{Roadmap.}
The remainder of this section unfolds as follows.
Section~\ref{sec:objective} establishes notation and formalizes the calibration objective: we seek an additive correction that systematically assigns higher scores to anomalies than to normal samples.
Section~\ref{sec:method_ctad} details the construction of~$\mP$ and~$\mQ$, and presents the OT-based calibration mechanism.
Section~\ref{sec:theory} provides rigorous theoretical analysis, proving that under mild separability conditions, anomalies receive systematically higher calibration terms than normals. This explains \emph{why} CTAD generalizes across diverse datasets.
Finally, Section~\ref{sec:algorithm} synthesizes these ideas into a complete algorithm and analyzes computational complexity, demonstrating that CTAD adds negligible overhead.

\subsection{Objective and Design Philosophy} 
\label{sec:objective}

In anomaly detection, the fundamental principle is that anomaly scores should reflect the degree of deviation from normality—higher scores indicate greater confidence that a sample is anomalous. This principle remains valid even when a detector does not perfectly model the normal distribution, as long as it assigns systematically higher scores to anomalies than to normal samples.

Building on this insight, we design a sample-specific calibration term $\Delta$ that amplifies the separation between anomalous and normal samples. Formally, we seek $\Delta$ satisfying:

\begin{equation}
\label{eq:calibration_objective}
\begin{aligned}
    \mathbb{E}[\Delta_{\mathcal{A}}] > \mathbb{E}[\Delta_{\mathcal{N}}], 
\end{aligned}
\end{equation}
where $\Delta_\mathcal{A}$ and $\Delta_\mathcal{N}$ denote calibration terms for anomalous and normal samples respectively. The calibrated score of $\mathbf{x}_{test}\in \mathcal{D}_{test}$ becomes $s_{test}^* \gets s_{test} + \Delta_{test}$, where $s_{test}$ is the original anomaly score from any base detector $f$.

The key challenge is: \textit{how can we construct such a calibration term without access to anomaly labels?} Our solution stems from a geometric intuition about distribution compatibility.

\subsection{Distribution Compatibility via Optimal Transport}
\label{sec:method_ctad}

\subsubsection{The Two-Distribution Framework}

We characterize normality through two complementary representations of $\mathcal{D}_{train}$:

\noindent\textbf{Empirical snapshot.} We construct an empirical distribution by randomly sampling $M$ points from $\mathcal{D}_{train}$:
\begin{equation}
\label{eq:original_p}
\begin{aligned}
\mP = \frac{1}{M} \sum_{i=1}^{M} \delta _{\mathbf{x}_{i}},
\end{aligned}
\end{equation}
where $\delta _{\mathbf{x}_{i}}$ denotes the Dirac measure at $\mathbf{x}_i$. This provides a sample-based approximation of the training distribution.

\noindent\textbf{Structural representation.} We capture the global structure through K-means clustering applied to $\mathcal{D}_{train}$:
\begin{equation}
\label{eq:kmeans}
\begin{aligned}
    \min_{C\in \mathbb{R}^{K\times D}} \frac{1}{N}\sum_{i=1}^{N} \min_{\tilde{y_i}\in \{0,1\}^K} \|\mathbf{x}_i - \tilde{y_i}^\mathrm{T} C\|^2,\  \text{ s.t. } \tilde{y_i}^\mathrm{T} \textbf{1}_K = 1, 
\end{aligned}
\end{equation}
where $\textbf{1}_K \in \mathbb{R}^K$ is a vector of ones, $\tilde{y_i}$ indicates cluster assignment, and $C$ is the centroid matrix. The centroids $\{C_j\}_{j=1}^K$ represent the modes of the normal distribution, which we encode as:
\begin{equation}
\label{eq:target_q}
\begin{aligned}
\mQ = \frac{1}{K}\sum_{j=1}^{K} \delta _{C_j}.
\end{aligned}
\end{equation}

Since both $\mP$ and $\mQ$ characterize the same underlying normal distribution, they should exhibit strong compatibility. This compatibility forms the basis of our approach.

\subsubsection{Measuring Distribution Disruption}

Given a test sample $\mathbf{x}_{test} \in \mathcal{D}_{test}$, we embed it into the empirical distribution:
\begin{equation}
\label{eq:extend_p}
\begin{aligned}
\mP = \frac{1}{M+1} \left(\sum_{i=1}^{M} \delta _{\mathbf{x}_{i}} + \delta_{\mathbf{x}_{test}}\right).
\end{aligned}
\end{equation}

\noindent The key insight is geometric: if $\mathbf{x}_{test}$ is normal, adding it to $\mP$ should preserve the compatibility with $\mQ$; if $\mathbf{x}_{test}$ is anomalous, it disrupts this compatibility. We quantify this disruption using Optimal Transport (OT):

\begin{equation}
\label{eq:ot_distance}
\begin{aligned}
\mathbf{OT}(\mP, \mQ) &= \langle\mathbf{T^*},\mathbf{C}\rangle = \sum_{i=1}^{M+1}\sum_{j=1}^{K}\mathbf{T}_{ij}^*\mathbf{C}_{ij},\\
\text{s.t.}\quad  \mathbf{T^{\ast}}&=\underset{\mathbf{T}\in\Pi(\mP,\mQ)}{\operatorname*{\arg\min}}\langle\mathbf{T},\mathbf{C}\rangle, \\
\Pi(\mP,\mQ) &:= \left\{\textbf{T} \,\big|\, \sum_{i=1}^{M+1} \mathbf{T}_{ij}=\frac{1}{K},  \sum_{j=1}^{K} \mathbf{T}_{ij}=\frac{1}{M+1}\right\},
\end{aligned}
\end{equation}
where $\langle \cdot, \cdot \rangle$ is the Frobenius inner product, $\mathbf{T}^* \in \mathbb{R}^{(M+1)\times K}_{\geq 0}$ is the optimal transport plan, and $\mathbf{C}\in \mathbb{R}^{(M+1)\times K}$ is the cost matrix with $\mathbf{C}_{ij} = \|\mathbf{x}_i - C_j\|$. See Section~\ref{sec:ot} for OT background.

\noindent\textbf{Geometric interpretation.} The transport plan $\mathbf{T}^*$ specifies how to optimally match points in $\mP$ to centroids in $\mQ$. The marginal constraints ensure mass conservation: each point in $\mP$ must distribute exactly $\frac{1}{M+1}$ mass to the centroids. Critically, the test sample $\mathbf{x}_{test}$ (corresponding to row $M+1$ in $\mathbf{T}^*$) cannot escape this obligation—it must transport its mass regardless of its position. An anomaly far from all centroids pays a high transport cost, while a normal sample near centroids pays little, where the theoretical analysis is supported in Section~\ref{sec:theory}.

\subsection{Theoretical Foundation}
\label{sec:theory}

We now establish that OT distance provides the desired separation between anomalies and normal samples.

\vspace{0.5em}
\noindent
\textsc{\textbf{Proposition 1}} (Lower Bound).
\textit{
Given $\mP = \frac{1}{M+1} (\sum_{i=1}^{M} \delta _{\mathbf{x}_{i}} + \delta_{\mathbf{x}_{test}})$ and $\mQ = \frac{1}{K}\sum_{j=1}^{K} \delta _{C_j}$, the optimal transport distance satisfies:
\begin{equation}
\label{eq:lower_bound}
\mathbf{OT}(\mP, \mQ) \geq \frac{1}{M+1} \|\mathbf{x}_{test}-C_{j^*}\|,
\end{equation}
where $C_{j^*} =\underset{C_j\in C}{\operatorname*{\arg\min}}\|\mathbf{x}_{test}-C_j\|$ is the nearest centroid to $\mathbf{x}_{test}$.
}

\vspace{0.5em}
\noindent
\textit{\textbf{Proof.}} We decompose the OT distance into contributions from reference samples (Eq.~\ref{eq:original_p}) and the test sample:
\begin{equation}
\begin{aligned}
    \mathbf{OT}(\mP,\mQ) = \underbrace{\sum_{i=1}^{M}\sum_{j=1}^{K}\mathbf{T}_{ij}^*\mathbf{C}_{ij}}_\text{for reference samples} + \underbrace{\sum_{j=1}^K\mathbf{T}_{M+1, j}^*\mathbf{C}_{M+1, j}}_\text{for test sample}.
\end{aligned}
\end{equation}
Since all terms are non-negative, we can lower bound by considering only the test sample's contribution:
\begin{equation}
\begin{aligned}
    \mathbf{OT}(\mP,\mQ) \geq \sum_{j=1}^K\mathbf{T}_{M+1, j}^*\|\mathbf{x}_{test} - C_j\|.
\end{aligned}
\end{equation}
The marginal constraint requires $\sum_{j=1}^K\mathbf{T}_{M+1, j}^* = \frac{1}{M+1}$. Since $\mathbf{T}_{M+1,j}^* \geq 0$ forms a valid probability distribution over centroids, we apply the weighted average inequality:
\begin{equation}
\begin{aligned}
    \sum_{j=1}^K\mathbf{T}_{M+1, j}^*\|\mathbf{x}_{test} - C_j\| &\geq \left(\sum_{j=1}^K\mathbf{T}_{M+1, j}^*\right) \cdot \min_{j\in[K]}\|\mathbf{x}_{test} - C_j\|\\
    &= \frac{1}{M+1} \|\mathbf{x}_{test} - C_{j^*}\|. 
\end{aligned}
\end{equation}

\vspace{0.5em}
\noindent
\textbf{Why this bound matters.} Proposition 1 establishes that $\mathbf{OT}(\mP,\mQ)$ is directly tied to the test sample's distance from normal structure (represented by centroids). The bound is tight when $\mathbf{x}_{test}$ can send all its mass to its nearest centroid. Crucially, this lower bound is \textit{enforced by the marginal constraint}—the test sample cannot avoid participating in the transport.

\subsubsection{Upper Bound and Gap Analysis}

We complement the lower bound with an upper bound that considers all samples:

\vspace{0.5em}
\noindent
\textsc{\textbf{Proposition 2}} (Upper Bound).
\textit{
Under the same setup, the OT distance is bounded above by:
\begin{equation}
\label{eq:upper_bound}
\mathbf{OT}(\mP, \mQ) \leq \frac{1}{M+1}\sum_{i=1}^{M+1} \min_{j\in[K]} \|\mathbf{x}_i - C_j\|,
\end{equation}
where $\mathbf{x}_{M+1} \equiv \mathbf{x}_{test}$.
}

\vspace{0.5em}
\noindent
\textit{\textbf{Proof sketch.}} Consider the greedy transport plan where each source point sends all its mass to its nearest centroid. While this may not satisfy target marginals exactly, a feasible plan can be constructed with cost at most the greedy cost. Since the optimal plan minimizes cost, the result follows. (Full proof in Appendix~\ref{appendix:proof_upper_bound}.)

\vspace{0.5em}
Combining the bounds, we obtain:
\begin{equation}
\label{eq:combined_bounds}
\frac{1}{M+1}\|\mathbf{x}_{test}-C_{j^*}\| \leq \mathbf{OT}(\mP, \mQ) \leq \frac{1}{M+1}\left(\sum_{i=1}^{M} d_i + d^*\right),
\end{equation}
where $d_i = \min_{j\in[K]}\|\mathbf{x}_i - C_j\|$ for reference samples and $d^* = \|\mathbf{x}_{test} - C_{j^*}\|$ for the test sample.

\noindent\textbf{Implication.} Let $\bar{d} = \frac{1}{M}\sum_{i=1}^M d_i$ denote the average nearest-centroid distance for reference samples. Since these are drawn from $\mathcal{D}_{train}$ (normal data), $\bar{d}$ is approximately constant across different test samples. The variation in $\mathbf{OT}(\mP,\mQ)$ is thus driven by $d^*$, i.e., the test sample's distance from centroids ($\min_{j\in[K]} \|\mathbf{x}_{test} - C_j\|$).

\subsubsection{Expected Separation Between Anomalies and Normals}

To formalize when CTAD achieves the calibration objective (Eq.~\ref{eq:calibration_objective}), we introduce two natural assumptions:

\vspace{0.5em}
\noindent
\textsc{\textbf{Assumption 1}} (Normal Clustering).
\textit{Normal samples lie close to cluster centroids:}
\begin{equation}
\label{eq:assumption_normal}
\mathbb{E}_{\mathbf{x}_{test}\in \mathcal{N}}\left[\min_{j\in[K]} \|\mathbf{x}_{test} - C_j\|\right] \leq \epsilon,
\end{equation}
\textit{where $\mathcal{N}$ denotes normal samples in $\mathcal{D}_{test}$ and $\epsilon > 0$ is a small constant depending on clustering quality.}

This assumption is justified by the K-means objective (Eq.~\ref{eq:kmeans}): if K-means successfully clusters the training data, normal test samples from the same distribution will be close to some centroid.

\vspace{0.5em}
\noindent
\textsc{\textbf{Assumption 2}} (Anomaly Separation).
\textit{Anomalous samples are distant from all centroids:}
\begin{equation}
\label{eq:assumption_anomaly}
\mathbb{E}_{\mathbf{x}_{test}\in \mathcal{A}}\left[\min_{j\in[K]} \|\mathbf{x}_{test} - C_j\|\right] \geq \eta,
\end{equation}
\textit{where $\mathcal{A}$ denotes anomalies in $\mathcal{D}_{test}$ and $\eta > \epsilon$.}

This assumption captures what makes something an anomaly: it does not belong to normal clusters. If anomalies were close to normal centroids, they would be indistinguishable from normal samples.

\vspace{0.5em}
\noindent
\textsc{\textbf{Theorem 1}} (Expected OT Gap).
\textit{
Under Assumptions 1 and 2, the expected OT distance satisfies:
\begin{equation}
\label{eq:main_theorem}
\mathbb{E}_{\mathbf{x}_{test}\in \mathcal{A}}[\mathbf{OT}(\mP, \mQ)] - \mathbb{E}_{\mathbf{x}_{test}\in \mathcal{N}}[\mathbf{OT}(\mP, \mQ)] \geq \frac{\eta - \bar{\epsilon}}{M+1},
\end{equation}
where $\bar{\epsilon} = (M+1)\epsilon$ accounts for the reference samples' contribution.
}

\vspace{0.5em}
\noindent
\textit{\textbf{Proof.}} From Proposition 1, the expected OT for anomalies satisfies:
\begin{equation}
\mathbb{E}_{\mathbf{x}_{test}\in \mathcal{A}}[\mathbf{OT}(\mP, \mQ)] \geq \frac{1}{M+1}\mathbb{E}_{\mathbf{x}_{test}\in \mathcal{A}}\left[\|\mathbf{x}_{test}-C_{j^*}\|\right] \geq \frac{\eta}{M+1}.
\end{equation}
From Proposition 2 and Assumption 1, the expected OT for normals satisfies:
\begin{equation}
\begin{aligned}
\mathbb{E}_{\mathbf{x}_{test}\in \mathcal{N}}[\mathbf{OT}(\mP, \mQ)] &\leq \frac{1}{M+1}\left(M\epsilon + \mathbb{E}_{\mathbf{x}_{test}\in \mathcal{N}}[d^*]\right) \\
&\leq \frac{(M+1)\epsilon}{M+1} = \epsilon.
\end{aligned}
\end{equation}
Taking the difference yields the result. 

\vspace{0.5em}
\noindent
\textbf{Interpretation.} Theorem 1 guarantees that under reasonable separability conditions, anomalies systematically receive higher OT scores than normal samples. The gap is positive when $\eta > (M+1)\epsilon$, which requires anomalies to be sufficiently separated from normal structure. The factor $(M+1)$ reflects the averaging effect: larger $M$ requires stronger separation but also reduces variance. (See Appendix~\ref{appendix:variance_analysis} for variance analysis.)

\noindent
\textbf{Discussion: Why OT over simpler distance measures?}
A natural question arises: since the separation condition $\eta > (M+1)\epsilon$ suggests that anomalies lie farther from centroids, why not simply use the nearest centroid distance $\min_j \|\mathbf{x}_{test} - C_j\|$ as the calibration signal? While such a measure would achieve a theoretical gap of $\eta - \epsilon$ under the same assumptions, OT provides three essential advantages. \textit{First}, OT captures global structural compatibility between distributions $\mP$ and $\mQ$ rather than local point-wise proximity; a test sample equidistant to multiple centroids may still disrupt the optimal alignment, a nuance that nearest-centroid distance misses. \textit{Second}, OT is robust to clustering uncertainty by integrating information from all $M$ reference samples and $K$ centroids through the optimal transport plan, mitigating the impact of suboptimal clustering. \textit{Third}, the marginal constraints in OT enforce mandatory mass transportation, systematically amplifying costs for anomalies that deviate from the normal manifold. Empirically, OT consistently outperforms nearest-centroid calibration across diverse datasets (see Fig.~\ref{fig:alternative_calibration} in Section~\ref{sec:comparison_methods} for detailed comparisons).

\subsection{Framework Overview and Discussion}
\label{sec:algorithm}

\textbf{Framework overview.} Armed with theoretical guarantees, we construct the sample-specific calibration term:
\begin{equation}
\label{eq:calibration_formula}
\begin{aligned}
\Delta_{test}& = \mathbf{OT}(\mP, \mQ), \\
s_{test}^* &\gets s_{test} + \lambda*\Delta_{test},
\end{aligned}
\end{equation}
where $s_{test}$ is the original anomaly score from any base detector $f(\mathbf{x}_{test})$, $\lambda\in \mathbb{R}$ is the hyperparameter controlling the calibration weight. By Theorem 1, we have $\mathbb{E}[\Delta_a] > \mathbb{E}[\Delta_n]$ under Assumptions 1-2, satisfying the calibration objective (Eq.~\ref{eq:calibration_objective}).
Algorithm~\ref{algorithm_ctad} in Appendix~\ref{appendix:ctad_details} summarizes the complete CTAD workflow. The offline phase computes K-means centroids once; the online phase performs fast OT computation for each test sample.

\noindent\textbf{Computational complexity.} K-means runs once with $O(NKD)$ cost per iteration. For each test sample, computing the cost matrix requires $O((M+1)KD)$, and solving the linear program (OT) requires $O((M+1)^3 K)$ via network simplex. With typical values $M=20$, $K=5$, the per-sample cost is negligible. (See Section~\ref{sec:computation} for detailed analysis.)

\noindent\textbf{Design choices and alternatives.} Our framework makes several principled design decisions:

\begin{itemize}[leftmargin=*, noitemsep, topsep=3pt]
\item \textit{Why embed $\mathbf{x}_{test}$ in $\mP$?} This design ensures the test sample participates in transport via marginal constraints, enabling tight theoretical bounds. 
% Alternative designs (e.g., comparing $\mP$ vs. $\mP \cup \{\mathbf{x}_{test}\}$) lack clean guarantees.

\item \textit{Why K-means?} Centroids minimize within-cluster variance and efficiently capture distribution modes. More complex methods (GMM, spectral clustering) could be used but add computational cost without clear benefit in our experiments.

% \item \textit{Why uniform weights?} Equal weighting simplifies analysis and avoids hyperparameter tuning. Non-uniform weights (e.g., cluster-size-proportional) could be explored but are not necessary for good performance.

% \item \textit{Why 1-Wasserstein?} Linear cost sensitivity strikes the right balance—moderate anomalies aren't over-penalized, and extreme anomalies aren't under-penalized. The 2-Wasserstein uses squared costs, which can be overly sensitive.

\item \textit{Why Euclidean distance cost?} We use Euclidean distance $\|\mathbf{x}_i - C_j\|$ as the cost function in the transport problem (Eq.~\ref{eq:ot_distance}), which provides linear cost sensitivity. This strikes the right balance—moderate anomalies aren't over-penalized (as they would be with squared costs), and extreme anomalies aren't under-penalized.
\end{itemize}

% \noindent\textbf{When CTAD may struggle.} Our method relies on geometric separation (Assumptions 1-2). Failure modes include:
% (1) \textit{Camouflaged anomalies:} If anomalies lie near centroids ($\eta \approx \epsilon$), they appear normal. 
% (2) \textit{Poor clustering:} If K-means fails to capture structure (wrong $K$, non-spherical clusters), centroids misrepresent normality.
% (3) \textit{Distribution shift:} If test normals differ from training normals, they may appear anomalous.
% We address these limitations through careful hyperparameter selection and validate robustness in Section~\ref{sec:experiments}.

%% file: 5_experiments.tex
\begin{table*}[t!]
\footnotesize
\centering
\captionsetup{font=small}
\captionsetup{skip=0pt}
\caption{Statistical summary of detection performance improvement achieved by CTAD over base detectors ($f$) on 34 datasets. ``Baseline'' indicates the average performance of the base detector ($f$) on all datasets. ``Improv. ($\Delta$)'' shows the average absolute performance improvement by CTAD over $f$. ``Improv. (\%)'' indicates the average relative performance improvement. ``Win Count'' represents the number of datasets where CTAD improved over $f$. $p$-value measures the significance of improvements across 34 datasets via paired one-tailed t-test (values $< 0.05$ indicate statistical significance).}
\label{table:main summary}
% \resizebox{\linewidth}{!}{
\setlength{\tabcolsep}{3.0mm}{
\begin{tabular}{c|ccccc|ccccc}
\toprule
\multirow{2}{*}{Detector} & \multicolumn{5}{c|}{AUC-PR} & \multicolumn{5}{c}{AUC-ROC}        \\ \cline{2-6} \cline{7-11}
         & Baseline & Improv. ($\Delta$) & Improv. (\%) & Win Count  & $p$-value       & Baseline & Improv. ($\Delta$) & Improv. (\%) & Win Count  & $p$-value \\ \midrule
PCA      & .5942    & .0515       & 48.89            & 22 / 34  & .0244         & .7316    & .0509       & 14.47            & 23 / 34  & .0054 \\
IForest  & .5664    & .0528       & 15.62            & 25 / 34  & .0003         & .7533    & .0202       & 3.28             & 25 / 34  & .0082 \\
OCSVM    & .5228    & .0550       & 39.32            & 23 / 34  & .0295         & .7111    & .0443       & 9.78             & 24 / 34  & .0243 \\
ECOD     & .5344    & .0542       & 19.72            & 23 / 34  & .0094         & .7047    & .0461       & 7.77             & 25 / 34  & .0023 \\
KNN      & .6414    & .0255       & 8.29             & 34 / 34  & 6.65e-06      & .7661    & .0271       & 7.39             & 34 / 34  & 9.87e-06 \\
MCM      & .6260    & .0578       & 15.80            & 25 / 34  & .0089         & .7666    & .0487       & 10.12            & 25 / 34  & .0075 \\
DRL      & .7290    & .0237       & 4.17             & 18 / 34  & .0012         & .8518    & .0083       & 1.18             & 18 / 34  & .0067 \\
\bottomrule
\end{tabular}
}
\end{table*}

\section{Experiments \& Analysis} \label{sec:experiments}
We conduct comprehensive experiments to validate CTAD's effectiveness and generalizability. Our evaluation addresses the following research questions: \textbf{(RQ1)} Does CTAD consistently improve existing TAD models across diverse datasets? \textbf{(RQ2)} How does CTAD compare to alternative calibration strategies? \textbf{(RQ3)} How sensitive is CTAD to its hyperparameters? \textbf{(RQ4)} Does the empirical evidence support our theoretical foundations (i.e., do anomalies exhibit higher OT distances than normal samples as claimed in Theorem 1)? \textbf{(RQ5)} What is the computational overhead of CTAD?

\subsection{Experimental Setup}
\textbf{Datasets and Metrics.} 
% As described before, AD on tabular datasets are challenging due to the heterogeneity, complexity, and diversity of tabular data.
% To validate the effectiveness of the proposed method, we select 34 datasets from Outlier Detection DataSets (OODS)~\cite{rayana2016odds} and Anomaly Detection Benchmark (ADBench)~\cite{han2022adbench}.
% Table~\ref{table_dataset_properties} shows the statistics of the 34 heterogeneous tabular datasets that are included for a comprehensive evaluation.
% Following previous works~\cite{yin2024mcm, shenkar2022anomaly}, we randomly partition the normal data of each dataset into two equal halves. The training dataset consists of one-half of the normal data, while the testing dataset comprises the other half of the normal data combined with all the abnormal instances. We employ Area Under the Precision-Recall Curve (AUC-PR) and Area Under the Receiver Operating Characteristic Curve (AUC-ROC) as our evaluation metrics.
Tabular anomaly detection presents unique challenges due to the heterogeneity, complexity, and diversity of tabular data structures. To comprehensively evaluate CTAD's effectiveness, we select 34 datasets from two established benchmarks: Outlier Detection DataSets (OODS)~\cite{rayana2016odds} and Anomaly Detection Benchmark (ADBench)~\cite{han2022adbench}.
Table~\ref{table_dataset_properties} of Appendix~\ref{appendix:dataset} presents the statistics of these datasets, which vary significantly in dimensionality (5 to 768 features), sample size (80 to 49,097 instances), and anomaly ratio (1\% to 75.4\%), providing a rigorous testbed for evaluating model-agnostic calibration.
Following established protocols~\cite{yin2024mcm, ye2025drl}, we randomly partition the normal data of each dataset into two equal halves. The training set consists of one half of the normal data, while the test set comprises the other half of normal data combined with all anomalous instances. We employ Area Under the Precision-Recall Curve (AUC-PR) and Area Under the Receiver Operating Characteristic Curve (AUC-ROC) as our evaluation metrics.

\textbf{Baseline Models.}
To demonstrate CTAD's model-agnostic nature and comprehensive applicability, we carefully select 7 representative TAD algorithms that span \textit{all four major categories} identified in Section~\ref{sec:related_work} (Related Work), encompassing both classical methods and recent deep learning approaches:
\textbf{\textit{Density Estimation-based}}: KNN~\citep{ramaswamy2000efficient} (distance-based density estimation) and ECOD~\cite{li2022ecod} (empirical cumulative distribution); 
\textbf{\textit{Classification-based}}: OCSVM~\cite{scholkopf1999support} (kernel-based one-class classification); 
\textbf{\textit{Isolation-based}}: IForest~\cite{liu2008isolation} (isolation forest);
\textbf{\textit{Reconstruction-based}}: PCA~\cite{shyu2003novel} (classical linear reconstruction), MCM~\citep{yin2024mcm} (deep masked cell modeling), and DRL~\cite{ye2025drl} (decomposed representation learning with orthogonal basis).
This comprehensive selection ensures our evaluation covers the full spectrum of TAD paradigms, from traditional statistical methods (PCA, IForest, OCSVM) to modern deep learning techniques (MCM, DRL), validating CTAD's generalizability across diverse architectural assumptions and detection principles. Detailed descriptions of each baseline are provided in Appendix~\ref{appendix:baselines}.
% \begin{itemize}[leftmargin=*]
% \vspace{-0.5em}
%     \item \textit{Density Estimation-based}: KNN~\citep{ramaswamy2000efficient} (distance-based density estimation) and ECOD~\cite{li2022ecod} (empirical cumulative distribution). These methods detect anomalies by measuring deviation from estimated normal density.

%     \item \textit{Classification-based}: OCSVM~\cite{scholkopf1999support} (kernel-based one-class classification). This method learns a decision boundary that separates normal samples from the origin in a high-dimensional space.

%     \item \textit{Reconstruction-based}: PCA~\cite{shyu2003novel} (classical linear reconstruction), MCM~\citep{yin2024mcm} (deep masked cell modeling), and DRL~\cite{ye2025drl} (deep representation learning with orthogonal basis). These methods identify anomalies through reconstruction error.

%     \item \textit{Isolation-based}: IForest~\cite{liu2008isolation} (isolation forest). This method assumes anomalies are easily isolated and detects them through path length in random trees.
% \vspace{-0.5em}
% \end{itemize}

\textbf{Implementation Details.}
% The CTAD architecture remains consistent across all datasets.
% Specifically, the size $M$ for random sampling is set to 20 by default and the size $K$ of centroids for K-Means clustering is set to 5. For each dataset, the calculated $\{\Delta_{test}\}$ over test set is consistent across all well-trained TAD models. Therefore, CTAD is model-agnostic and can be used to calibrate any well-trained TAD models.
CTAD maintains a consistent configuration across all experiments to ensure fair comparison. We set the number of reference samples $M = 20$, the number of K-means centroids $K = 5$ and the calibration weight $\lambda = 1.0$ as default values, which we find provide a good balance between calibration quality and computational efficiency (see sensitivity analysis in Section~\ref{sec:sensitivity}). Importantly, for each dataset, the computed calibration term $\{\Delta_{test}\}$ over the test set is identical across all base TAD models, demonstrating CTAD's true model-agnostic property---the calibration mechanism operates independently of the underlying detector architecture.

\subsection{Main Results (RQ1)}
\label{sec:main_results}

\textbf{Improvement over Base Detectors.}
% We train the 5 different mainstream TAD models as well as these models coupled with CTAD on all 34 tabular datasets.
% Note that this is a large-scale experiment that result in 5(models)$\times$34(datasets)$\times$2(metrics)$\times$2(w/o \& w/ CTAD) = 680 numerical results.
% We cannot fully display these results here due to space limitations.
% Therefore, we provide a summarization of the experimental results in Table \ref{table:main summary}, with the full results illustrated in Table~\ref{appendix:full_aucpr} and Table~\ref{appendix:full_aucroc} in Appendix~\ref{appendix:results}. We can observe that:
We conduct a large-scale experimental study training each of the 7 TAD models with and without CTAD on all 34 datasets, resulting in 7 (models) $\times$ 34 (datasets) $\times$ 2 (metrics) $\times$ 2 (w/o \& w/ CTAD) = 952 numerical results.
Due to space constraints, Table~\ref{table:main summary} presents statistical summaries of the improvements achieved by CTAD, while complete per-dataset results are provided in Tables~\ref{appendix:full_aucpr} and~\ref{appendix:full_aucroc} of Appendix~\ref{appendix:full_improv}.

From the results, we make the following key observations:
\textit{\textbf{(1) Universal improvement across detectors.}} CTAD consistently enhances all source detectors across both AUC-PR and AUC-ROC metrics.
Notably, CTAD improves KNN on all 34/34 datasets for both metrics, demonstrating exceptional consistency.
For other detectors, CTAD achieves improvements on 18-25 out of 34 datasets (53\%-74\% win rate), with average relative improvements ranging from 1.18\% to 48.89\%.
We conduct paired one-tailed t-tests (with $\alpha=0.05$) comparing each TAD model with its CTAD-enhanced version across all 34 datasets.
Remarkably, improvements are statistically significant at the 95\% confidence level for all detectors (all $p$-values $< 0.05$), with KNN achieving particularly strong significance ($p < 10^{-5}$).
This demonstrates CTAD's superior adaptability and generalizability across different model architectures and dataset characteristics.
\textit{\textbf{(2) No universal TAD winner.}}
As shown in Tables~\ref{appendix:full_aucpr} and~\ref{appendix:full_aucroc}, there is no single TAD model that consistently outperforms others across all datasets---the "universal winner" does not exist. This is consistent with our claims and prior findings~\cite{han2022adbench}. This observation reinforces the importance of model-agnostic calibration frameworks like CTAD, which can enhance any detector rather than seeking a single optimal model.
\textit{\textbf{(3) Improvement even on strong baselines.}}
CTAD achieves meaningful improvements even on the best-performing detector (DRL). Despite DRL's strong baseline performance, CTAD still manages an average relative gain of 4.17\%/1.18\% for AUC-PR/AUC-ROC respectively. This demonstrates that CTAD's distribution compatibility perspective provides complementary information beyond what sophisticated deep learning models learn from anomaly patterns alone. Even when base detectors capture complex feature relationships, the OT-based calibration adds value by measuring how well test samples align with the global structure of the normal distribution.

\begin{figure}[!t]
\centering
\includegraphics[width=0.95\linewidth]{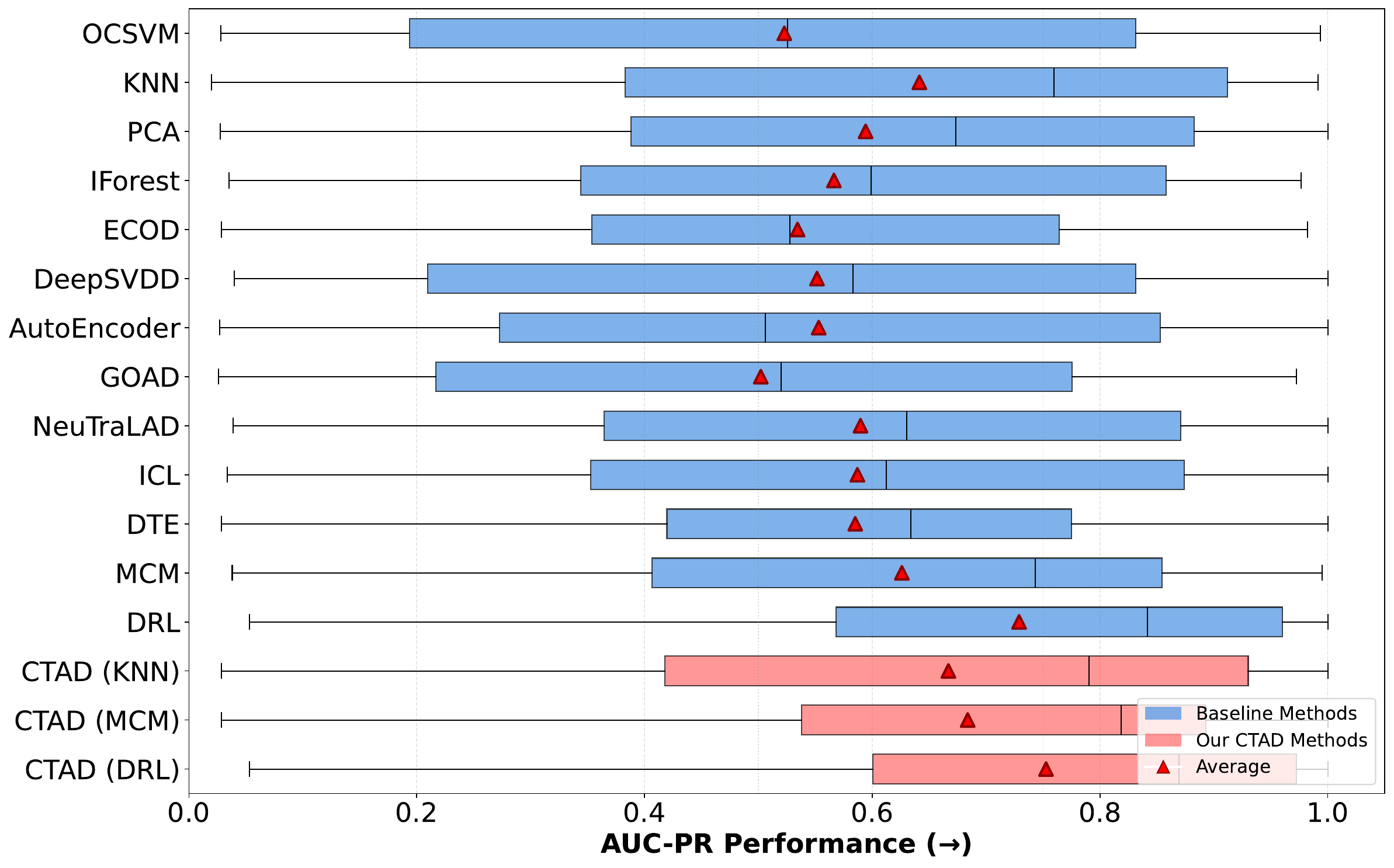}
\captionsetup{font=small}
\captionsetup{skip=0pt}
\caption{Comparison of all models' performance across different datasets (AUC-PR). The red triangles represent the average value. CTAD with DRL as base detector achieves the SOTA performance.} 
\label{fig:overall_aucpr}
% \vspace{-1em}
\end{figure}

\textbf{Comparison with Other Baseline Methods.}
Beyond comparing CTAD-enhanced models against their base versions, we also position CTAD within the broader landscape of tabular anomaly detection methods.
As shown in Fig.~\ref{fig:overall_aucpr}, when coupled with DRL, CTAD achieves the highest average AUC-PR performance across all 34 datasets, surpassing both classical methods (PCA, IForest, OCSVM, ECOD, KNN) and recent deep learning approaches (MCM, DRL without calibration).
This demonstrates that CTAD's post-processing calibration framework, when applied to strong base detectors, can outperform standalone state-of-the-art methods.
Complete per-dataset comparisons are provided in Tables~\ref{appendix:methodsPR} and~\ref{appendix:methodsROC} of Appendix~\ref{appendix:full_comparison}.

\subsection{Comparison with Alternative Calibration Methods (RQ2)}
\label{sec:comparison_methods}

\begin{figure}[!t]
\centering
\includegraphics[width=\linewidth]{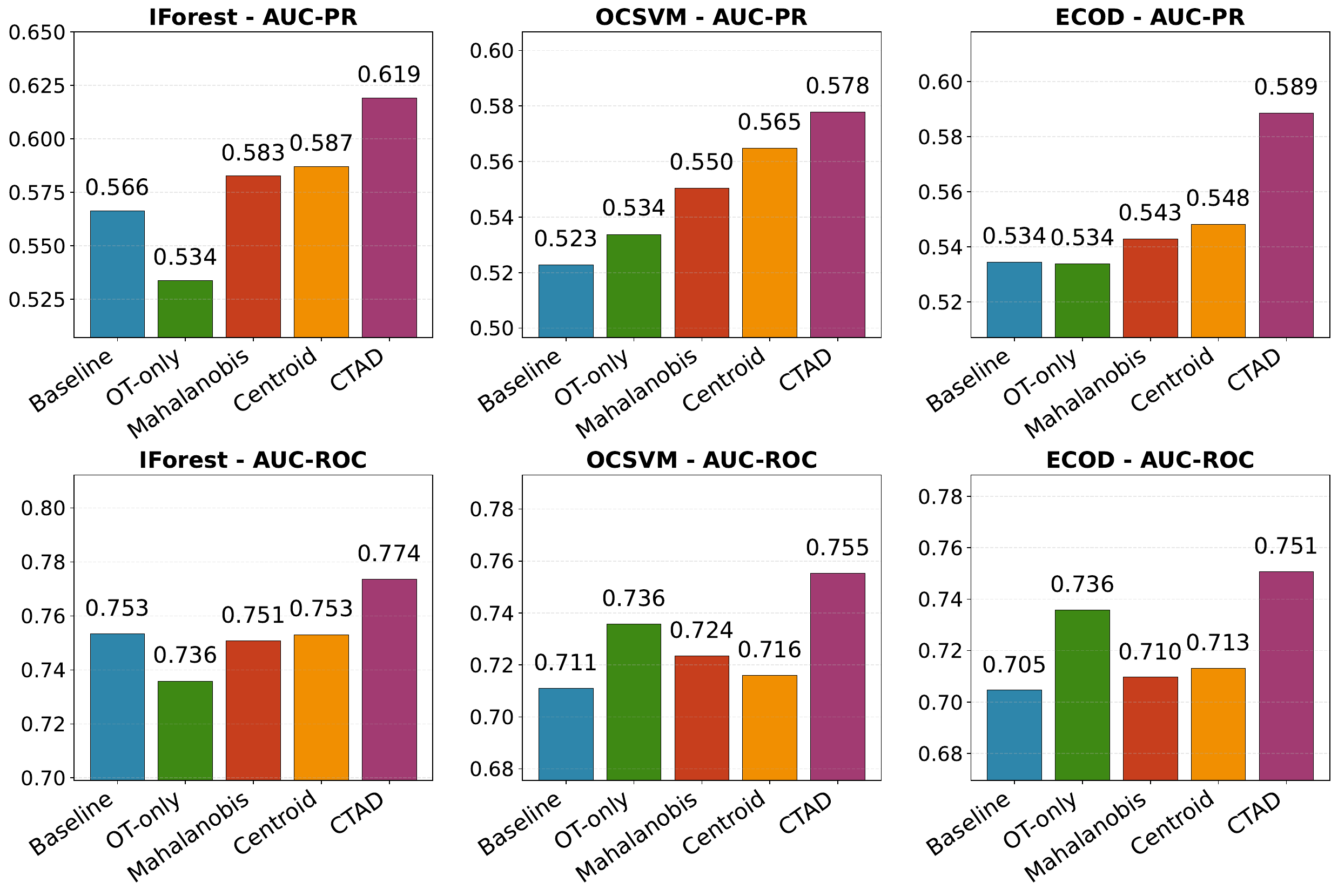}
\captionsetup{font=small}
\captionsetup{skip=0pt}
\caption{Comparison with Alternative Calibration Methods across 34 datasets. CTAD consistently outperforms alternative calibration strategies including centroid distance, Mahalanobis distance, and OT-only approaches, demonstrating the effectiveness of our optimal transport-based calibration framework.} 
\label{fig:alternative_calibration}
% \vspace{-0.5em}
\end{figure}

To demonstrate CTAD's effectiveness against alternative calibration strategies, we compare five approaches across 3 representative detectors (IForest, OCSVM, ECOD) on all 34 datasets, where $s_{test}$ is the original anomaly score from any base detector $f(\mathbf{x}_{test})$:
\textit{Baseline}: Original detector scores without calibration, $s_{test}^* = s_{test}$;
\textit{OT-only}: Pure OT distance ignoring base detector, $s_{test}^* = \mathbf{OT}(\mP, \mQ)$;
\textit{Centroid Distance}~\cite{de2000mahalanobis}: Minimum distance to K-means centroids, $s_{test}^* = s_{test} + \lambda \cdot \min_j \|\mathbf{x}_{test} - C_j\|$;
\textit{Mahalanobis}: Mahalanobis distance to training mean, which accounts for feature correlations via covariance matrix: $s_{test}^* = s_{test} + \lambda \cdot \sqrt{(\mathbf{x}_{test} - \boldsymbol{\mu})^\top \boldsymbol{\Sigma}^{-1} (\mathbf{x}_{test} - \boldsymbol{\mu})}$, where $\boldsymbol{\mu}$ and $\boldsymbol{\Sigma}$ are the mean and covariance of training data;
\textit{CTAD (Ours)}: OT-based calibration, $s_{test}^* = s_{test} + \lambda \cdot \mathbf{OT}(\mP, \mQ)$.
% \begin{itemize}[leftmargin=*]
% \vspace{-0.5em}
%     \item \textbf{Baseline}: Original detector scores without calibration, $s_{test}^* = s_{test}$
%     \item \textbf{CTAD (Ours)}: OT-based calibration, $s_{test}^* = s_{test} + \lambda \cdot \mathbf{OT}(\mP, \mQ)$
%     \item \textbf{Centroid Distance}: Minimum distance to K-means centroids, $s_{test}^* = s_{test} + \lambda \cdot \min_j \|\mathbf{x}_{test} - C_j\|$
%     \item \textbf{Mahalanobis}: Mahalanobis distance to training mean, which accounts for feature correlations via covariance matrix: $s_{test}^* = s_{test} + \lambda \cdot \sqrt{(\mathbf{x}_{test} - \boldsymbol{\mu})^\top \boldsymbol{\Sigma}^{-1} (\mathbf{x}_{test} - \boldsymbol{\mu})}$, where $\boldsymbol{\mu}$ and $\boldsymbol{\Sigma}$ are the mean and covariance of training data
%     \item \textbf{OT-only}: Pure OT distance ignoring base detector, $s_{test}^* = \mathbf{OT}(\mP, \mQ)$
% \vspace{-0.5em}
% \end{itemize}
Figure~\ref{fig:alternative_calibration} presents the averaged results across all datasets.

% \begin{figure*}[!t]
% \centering
% \includegraphics[width=\linewidth,height=8cm]{figures/calibration_methods_comparison.pdf}
% \caption{Comparison with Alternative Calibration Methods across 34 datasets. CTAD consistently outperforms alternative calibration strategies including centroid distance, Mahalanobis distance, and OT-only approaches, demonstrating the effectiveness of our optimal transport-based calibration framework.} 
% \label{fig:alternative_calibration}
% \end{figure*}

We observe that:
\textit{\textbf{(1) CTAD achieves best overall performance.}} CTAD consistently outperforms all alternative calibration methods across both AUC-ROC and AUC-PR metrics for all three base detectors. The improvements demonstrate that OT distance provides a principled and effective calibration signal compared to simpler geometric measures.
\textit{\textbf{(2) Centroid Distance is competitive but less principled.}} While centroid-based calibration performs reasonably well, it only considers the nearest centroid distance—a local geometric measure that ignores the global distributional structure. In contrast, CTAD's OT formulation provides a principled distance metric with theoretical guarantees (Theorem 1), considering the optimal alignment between the full empirical distribution (centroids + test sample) and reference samples from the training set.
\textit{\textbf{(3) Mahalanobis Distance underperforms.}} The global Mahalanobis distance shows limited improvements, particularly on high-dimensional datasets where covariance estimation becomes unreliable. Its assumption of a single Gaussian distribution is overly restrictive for heterogeneous tabular data.
\textit{\textbf{(4) OT-only validates the OT signal but combination is superior.}} Using OT distance alone (without base detector scores) performs competitively, confirming our theoretical result (Theorem 1) that OT distance provides a valid anomaly signal. However, the hybrid approach (CTAD) consistently outperforms OT-only, demonstrating that combining base detector signals with OT-based calibration yields superior performance. This validates our design philosophy: the calibration term complements rather than replaces the base detector, leveraging both learned anomaly patterns and distribution compatibility.
These results confirm that CTAD's OT-based calibration provides both theoretical soundness and practical effectiveness, striking the right balance between capturing distributional structure and leveraging learned anomaly patterns.

\subsection{Sensitivity Analysis (RQ3)}
\label{sec:sensitivity}

\begin{figure}[!t]
\centering
\includegraphics[width=\linewidth]{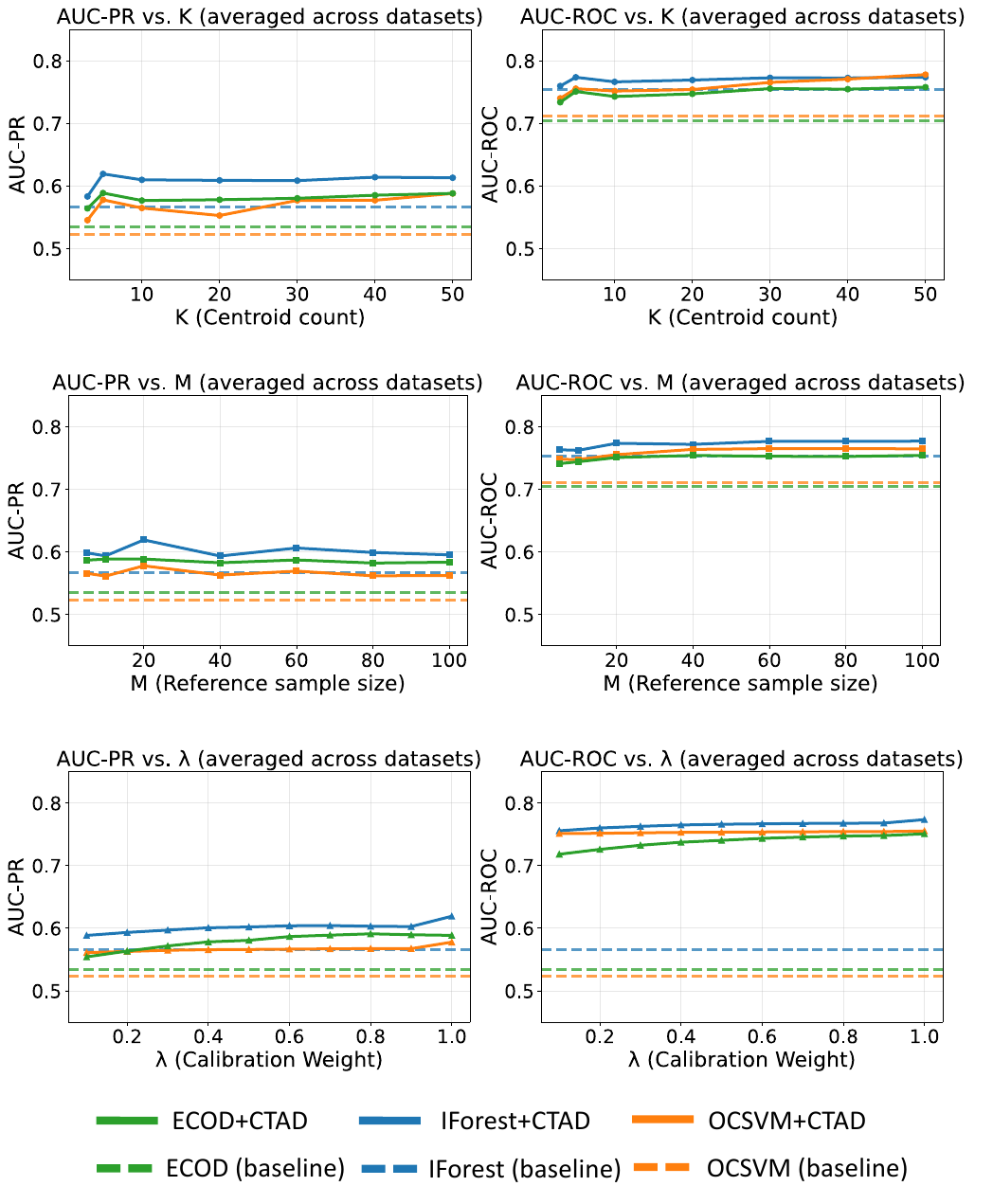}
\captionsetup{font=small}
\captionsetup{skip=0pt}
\caption{Hyperparameter ablation studies. Top: Effect of centroid count $K$. Middle: Effect of reference sample size $M$. Bottom: Effect of calibration weight $\lambda$.} 
\label{fig:sensitivity}
% \vspace{-1em}
\end{figure}

We conduct comprehensive ablation studies to understand the impact of three key hyperparameters: $K$ (number of centroids), $M$ (number of reference samples), and $\lambda$ (calibration weight), as illustrated in Fig.~\ref{fig:sensitivity}. The results are averaged over all datasets. We vary the number of centroids $K \in \{3, 5, 10, 20, 30, 40, 50\}$, the number of reference samples $M \in \{5, 10, 20, 40, 60, 80, 100\}$, and the calibration weight $\lambda \in \{0.1, 0.2, \ldots, 1.0\}$. The results demonstrate CTAD's remarkable robustness to hyperparameter choices: performance remains relatively stable across all tested ranges of $K$, $M$, and $\lambda$, with only minor fluctuations. The performance curves are largely flat, indicating that CTAD is not overly sensitive to specific hyperparameter values within the tested ranges. This robustness allows CTAD to be deployed with simple default hyperparameters ($K=5$, $M=20$, $\lambda=0.5$) without extensive dataset-specific tuning, making it highly practical for real-world applications where hyperparameter optimization may be infeasible.

% Results show that performance improves as $K$ increases from 3 to 10, then plateaus with slight degradation beyond $K=20$, suggesting $K \in [5, 15]$ as a robust range. For reference sample size, performance steadily improves with larger $M$ up to $M \approx 40$, with diminishing returns beyond $M=20$, making $M=20$ an optimal efficiency-effectiveness trade-off. The calibration weight $\lambda$ exhibits stable performance across $\lambda \in [0.3, 0.7]$, with peak performance around $\lambda \approx 0.5$, while too small values ($\lambda < 0.2$) underutilize the calibration signal and too large values ($\lambda > 0.8$) overpower the base detector. Overall, CTAD demonstrates robustness to hyperparameter choices within reasonable ranges, making it practical for real-world deployment without extensive tuning.

% \subsection{Experimental Evidence for OT(P,Q).}
% % \subsection{Experimental Evidence for $\mathbb{E}[\mathbf{OT}(\mP, \mQ)]$.}
% We also conduct experiments to validate that $\mathbb{E}_{\mathbf{x}_{test} \in \mathcal{A}}[\mathbf{OT}(\mP, \mQ)]$ has the potential to be higher than $\mathbb{E}_{\mathbf{x}_{test} \in \mathcal{N}}[\mathbf{OT}(\mP, \mQ)]$, as illustrated in Table~\ref{tab:ot_difference}.
% These statistics were computed on the test sets of all datasets. Notably, for the majority of datasets (31 out of 34), anomalies significantly increased the discrepancy between $\mP$ and $\mQ$ compared to normal samples. This observation further supports the use of this property to enable adaptive error correction in TAD models.

\begin{table}[t!]
\footnotesize
\centering
\captionsetup{font=small}
\captionsetup{skip=0pt}
\caption{
The difference between $\mathbb{E}_{\mathbf{x}_{test} \in \mathcal{N}}[\mathbf{OT}(\mP, \mQ)]$ and $\mathbb{E}_{\mathbf{x}_{test} \in \mathcal{A}}[\mathbf{OT}(\mP, \mQ)]$ across different datasets (full results are provided in Table~\ref{appendix:ot_difference} of Appendix~\ref{appendix:full_theory_evidence}).
The results are calculated based on $\mathcal{D}_{test}$.
``Increase (\%)'' indicates the relative improvement from $\mathbb{E}_{\mathbf{x}_{test} \in \mathcal{N}}[\mathbf{OT}(\mP, \mQ)]$ to $\mathbb{E}_{\mathbf{x}_{test} \in \mathcal{A}}[\mathbf{OT}(\mP, \mQ)]$, i.e., $\frac{(\mathbb{E}_{\mathbf{x}_{test} \in \mathcal{A}}[\mathbf{OT}(\mP, \mQ)] - \mathbb{E}_{\mathbf{x}_{test} \in \mathcal{N}}[\mathbf{OT}(\mP, \mQ)])}{\mathbb{E}_{\mathbf{x}_{test} \in \mathcal{N}}[\mathbf{OT}(\mP, \mQ)]}$. 
Anomalies exhibit substantially higher OT distances on 31/34 datasets (91.2\%).
}
\label{tab:ot_difference}
\setlength{\tabcolsep}{2.1mm}{
\begin{tabular}{ccccc}
\toprule
  & abalone   & amazon    & annthyroid & arrhythmia       \\ \midrule
$\mathbb{E}_{\mathbf{x}_{test} \in \mathcal{N}}[\mathbf{OT}(\mP, \mQ)]$ & 0.0179    & 0.0338    & 0.0007     & 1.4669     \\
$\mathbb{E}_{\mathbf{x}_{test} \in \mathcal{A}}[\mathbf{OT}(\mP, \mQ)]$ & 0.0512    & 0.035     & 0.0011     & 2.3242     \\
Increase (\%) & 185.3388  & 3.3587    & 50.4653    & 58.4472  \\ \bottomrule
\end{tabular}}
\vspace{-1em}
\end{table}

\subsection{Experimental Validation of Theoretical Foundations (RQ4)}

A key theoretical assumption underlying CTAD (formalized in Theorem 1) is that anomalies exhibit larger OT distances than normal samples: $\mathbb{E}_{\mathbf{x}_{test} \in \mathcal{A}}[\mathbf{OT}(\mP, \mQ)] > \mathbb{E}_{\mathbf{x}_{test} \in \mathcal{N}}[\mathbf{OT}(\mP, \mQ)]$. We empirically validate this foundational property across all datasets, providing crucial evidence that our theoretical framework holds in practice. Table~\ref{tab:ot_difference} presents representative results from 4 datasets (full results for all 34 datasets in Table~\ref{appendix:ot_difference} of Appendix~\ref{appendix:full_theory_evidence}). The results strongly support our theoretical assumption: for 31 out of 34 datasets (91.2\%), anomalies exhibit significantly higher average OT distances than normal samples, with a median relative increase of 61.7\% across all 34 datasets. Importantly, this OT gap property holds consistently across diverse dataset characteristics---varying dimensionality (5-768 features), sample sizes (80-49,097 instances), and anomaly ratios (1\%-75.4\%)---demonstrating the universality of distribution compatibility as a reliable calibration signal. These empirical findings validate Theorem 1's expected separation guarantee and provide strong evidence that our theoretical foundations translate to real-world data, explaining why CTAD's calibration mechanism systematically improves anomaly detection across diverse scenarios.

\subsection{Computational Complexity (RQ5)}
\label{sec:computation}
We analyze CTAD's runtime on 2 representative datasets (Cardio, Mnist) across 3 base detectors. Table~\ref{tab:runtime_analysis} presents per-sample inference times. On these tested configurations, CTAD's calibration overhead is very consistent within each dataset, demonstrating that the OT computation time is independent of the base detector's complexity. 
% For fast detectors like ECOD (0.003ms baseline on Cardio), the relative overhead appears large in percentage terms; however, for more computationally intensive detectors like OCSVM on high-dimensional data (1.056ms baseline on Mnist), CTAD adds only 17\% overhead. 
Importantly, CTAD's computational complexity is independent of the training set size during inference, as centroids are pre-computed offline. The online inference requires only $K$ OT computations per sample with complexity $O((M+1)^3 K)$, which with typical parameters ($M=20$, $K=5$) results in minimal cost. In these experiments, the absolute OT overhead remains under 0.2ms per sample across all configurations, making CTAD highly practical for real-world anomaly detection deployments where such overhead is negligible compared to the significant performance improvements achieved (1.18\%-48.89\% relative gains as shown in Section~\ref{sec:main_results}).

\begin{table}[t!]
\centering
\footnotesize
\captionsetup{font=small}
\captionsetup{skip=0pt}
\caption{Per-sample runtime analysis of CTAD overhead across 2 datasets and 3 detectors. Absolute OT time is negligible (<0.2ms per sample on average).}
\label{tab:runtime_analysis}
\setlength{\tabcolsep}{4.3mm}{
\begin{tabular}{lcccc}
\toprule
Dataset & Model & Baseline (ms) & OT Time (ms) \\
\midrule
cardio & IForest & 0.0143 & 0.1594 \\
cardio & OCSVM & 0.0149 & 0.1592 \\
cardio & ECOD & 0.0034 & 0.1591 \\
mnist & IForest & 0.0218 & 0.1768 \\
mnist & OCSVM & 1.0560 & 0.1751 \\
mnist & ECOD & 0.0409 & 0.1769 \\
% \midrule
% \multicolumn{2}{l}{\textit{Average}} & 0.1181 & 0.1699 & 2790.3 \\
% \multicolumn{2}{l}{\textit{Median}} & 0.0132 & 0.1672 & 1198.1 \\
% \multicolumn{2}{l}{\textit{Min}} & 0.0022 & 0.0351 & 16.6 \\
% \multicolumn{2}{l}{\textit{Max}} & 1.0560 & 0.4594 & 21014.8 \\
\bottomrule
\end{tabular}}
\end{table}

%% file: 6_conclusion.tex
\section{Conclusion}

In this paper, we addressed a fundamental challenge in tabular anomaly detection: existing methods often fail when their underlying assumptions are violated by the heterogeneity and diversity of real-world tabular data. Rather than proposing yet another assumption-based detector, we introduced CTAD (Calibrating Tabular Anomaly Detection), a model-agnostic post-processing framework that enhances any existing detector through optimal transport-based calibration. The core insight is that we characterize the normal distribution through two complementary representations—an empirical distribution from random sampling and a structural distribution from K-means centroids—and anomalies disrupt the compatibility between these views more than normal samples. We formalized this intuition with rigorous theoretical analysis (Theorem 1), proving that anomalies systematically receive higher calibration terms under mild separability assumptions, with tight probabilistic bounds.
Extensive experiments on 34 diverse tabular datasets with 7 detectors spanning all major TAD categories validate our approach: CTAD achieves statistically significant improvements across both classical and deep learning methods, while adding negligible computational overhead. The model-agnostic nature of CTAD offers practical advantages: it works as a drop-in enhancement without retraining, requires no dataset-specific tuning, and provides theoretical guarantees. We hope our work can facilitate more future research on tabular anomaly detection.

%% file: Appendix_final.tex
\clearpage
\appendix
\balance
\section{Additional Experimental Details}
\label{appendix:experiments}

\subsection{Datasets details}
\label{appendix:dataset}
Tabular anomaly detection presents unique challenges due to the heterogeneity, complexity, and diversity of tabular data structures. To comprehensively evaluate CTAD's effectiveness, we select 34 datasets from two established benchmarks: Outlier Detection DataSets (OODS)~\cite{rayana2016odds} and Anomaly Detection Benchmark (ADBench)~\cite{han2022adbench}.
Table~\ref{table_dataset_properties} presents the statistics of these datasets.
\begin{table}[h!]
\footnotesize
\centering
\caption{Dataset properties. We use 34 commonly used tabular anomaly detection datasets in this paper.}
\label{table_dataset_properties}
\begin{tabular}{lccc}
\toprule
                 & Samples & Dims & Anomalies  \\ \midrule
Abalone          & 4177    & 7    & 2081       \\
Amazon           & 10000   & 768  & 500        \\
Annthyroid       & 7200    & 6    & 534        \\
Arrhythmia       & 452     & 274  & 66         \\
Breastw          & 683     & 9    & 239        \\
% Campaign         & 41188   & 62   & 4640       \\
Cardio           & 1831    & 21   & 176        \\
Cardiotocography & 2114    & 21   & 466        \\
Comm.and.crime   & 1994    & 101  & 993        \\
Fault            & 1941    & 27   & 673        \\
Glass            & 214     & 9    & 9          \\
Hepatitis        & 80      & 19   & 13         \\
Imgseg           & 2310    & 18   & 990        \\
Ionosphere       & 351     & 33   & 126        \\
Lympho           & 148     & 18   & 6          \\
Mammography      & 11183   & 6    & 260        \\
Mnist            & 7603    & 100  & 700        \\
Musk             & 3062    & 166  & 97         \\
Optdigits        & 5216    & 64   & 150        \\
Parkinson        & 195     & 22   & 147        \\
Pendigits        & 6870    & 16   & 156        \\
Pima             & 768     & 8    & 268        \\
Satellite        & 6435    & 36   & 2036       \\
Satimage-2       & 5803    & 36   & 71         \\
Shuttle          & 49097   & 9    & 3511       \\
% SpamBase         & 4207    & 57   & 1679       \\
Speech           & 3686    & 400  & 61         \\
Thyroid          & 3772    & 6    & 93         \\
Vertebral        & 240     & 6    & 30         \\
Vowels           & 1456    & 12   & 50         \\
WDBC             & 367     & 30   & 10         \\
WPBC             & 198     & 33   & 47         \\
Wbc              & 378     & 30   & 21         \\
Wilt             & 4819    & 5    & 257        \\
Wine             & 129     & 13   & 10         \\
Yeast            & 1484    & 8    & 507        \\
\bottomrule
\end{tabular}
\end{table}
% \vspace{-1em}

\subsection{Baseline Model Details}
\label{appendix:baselines}

We compare CTAD against 7 representative anomaly detection methods spanning all major TAD categories and encompassing both classical and deep learning approaches:

\textbf{Density Estimation-based Methods:}
\begin{itemize}[leftmargin=*]
    \item \textbf{KNN}~\cite{ramaswamy2000efficient}: K-Nearest Neighbors. Detects anomalies based on distance to the $k$-th nearest neighbor. Samples with large distances to their neighbors are considered anomalous. This method captures local density deviation.

    \item \textbf{ECOD}~\cite{li2022ecod}: Empirical Cumulative distribution based Outlier Detection. Employs empirical cumulative distribution functions to estimate the tail probability of each feature, aggregating these probabilities to compute anomaly scores. ECOD is parameter-free and efficient for high-dimensional data.
\end{itemize}

\textbf{Classification-based Methods:}
\begin{itemize}[leftmargin=*]
    \item \textbf{OCSVM}~\cite{scholkopf1999support}: One-Class Support Vector Machine. Learns a decision boundary in a high-dimensional feature space using kernel methods, separating normal samples from the origin. Samples outside this boundary are flagged as anomalies.
\end{itemize}

\textbf{Reconstruction-based Methods:}
\begin{itemize}[leftmargin=*]
    \item \textbf{PCA}~\cite{shyu2003novel}: Principal Component Analysis. Projects data onto principal components and uses reconstruction error (difference between original and reconstructed data) as the anomaly score. High reconstruction error indicates deviation from normal patterns.

    \item \textbf{MCM}~\cite{yin2024mcm}: Masked Cell Modeling. A deep learning method that learns to reconstruct masked feature values based on unmasked features. Anomalies exhibit higher reconstruction errors as they deviate from learned normal patterns. MCM leverages masking strategies to capture inter-feature dependencies.

    \item \textbf{DRL}~\cite{ye2025drl}: Decomposed Representation Learning. Enforces a structured latent space where each normal representation is expressed as a linear combination of fixed orthogonal basis vectors. Anomalies fail to satisfy this structural constraint, resulting in higher anomaly scores. DRL represents the state-of-the-art in deep learning-based TAD.
\end{itemize}

\textbf{Isolation-based Methods:}
\begin{itemize}[leftmargin=*]
    \item \textbf{IForest}~\cite{liu2008isolation}: Isolation Forest. Builds an ensemble of isolation trees that recursively partition the feature space. Anomalies require fewer splits to isolate, resulting in shorter average path lengths. 
\end{itemize}

\textbf{Implementation details:}
\begin{itemize}[leftmargin=*]
    \item Classical methods (OCSVM, PCA, IForest, ECOD, KNN): Implemented using PyOD package~\cite{zhao2019pyod}
    \item MCM: Official implementation from original paper
    \item DRL: Official implementation from original paper
    \item Hyperparameters: Follow original papers for all methods to ensure fair comparison
    \item Identical preprocessing and dataset splits across all methods
    % \item Feature normalization: Z-score normalization applied to all datasets
\end{itemize}

Beyond comparing CTAD-enhanced models against their base versions, we also position CTAD within the broader landscape of tabular anomaly detection methods, including classical non-deep methods and recent deep learning based methods.
Specifically, OCSVM~\citep{scholkopf1999support}, KNN~\citep{ramaswamy2000efficient}, PCA~\citep{shyu2003novel}, IForest~\citep{liu2008isolation} and ECOD~\citep{li2022ecod} represent classic AD approaches that continue to maintain popularity.
In addition, we compare our method to recent deep learning based methods, namely Deep SVDD~\citep{ruff2018deep}, AutoEncoder~\citep{chen2018autoencoder}, GOAD~\citep{bergman2020classification}, NeuTraLAD~\citep{qiu2021neural}, ICL~\citep{shenkar2022anomaly}, DTE~\citep{livernoche2024on}, MCM~\citep{yin2024mcm} and DRL~\citep{ye2025drl}.
We use the popular PyOD python package~\citep{zhao2019pyod} to implement OCSVM, KNN, PCA, IForest, ECOD, Deep SVDD and AutoEncoder.
We use the DeepOD python library~\citep{xu2023deep} to implement GOAD, NeuTraLAD and ICL.
The implementation of the other methods is based on their official open-source code releases. 
Following latest works~\citep{yin2024mcm, ye2025drl}, We implement all baseline models' hyperparameters following their original papers.
All the methods are implemented with identical dataset partitioning and preprocessing procedures, following previous works~\citep{yin2024mcm,ye2025drl}. 

% \subsection{Hyperparameter Selection}

% \textbf{CTAD hyperparameters:}
% \begin{itemize}
%     \item $K$ (number of centroids): Selected via elbow method on training set
%     \begin{itemize}
%         \item Search range: $K \in \{3, 5, 7, 10, 15, 20\}$
%         \item Default: $K=5$ works well across most datasets
%     \end{itemize}
%     \item $M$ (reference samples): Fixed at $M=20$
%     \begin{itemize}
%         \item Balances variance reduction and computation
%         \item Ablation study shows diminishing returns beyond $M=20$
%     \end{itemize}
%     \item OT solver: Exact EMD via network simplex
%     \item Distance metric: Euclidean ($\ell_2$) distance
% \end{itemize}

% \textbf{Feature normalization:}
% All features are z-score normalized before both base detector training and CTAD calibration:
% \begin{equation}
% \mathbf{x}_{\text{norm}} = \frac{\mathbf{x} - \mathbb{E}[\mathbf{x}]}{\sqrt{\text{Var}[\mathbf{x}]}}
% \end{equation}
% This ensures OT distance is not dominated by high-variance features.

\subsection{CTAD Details}
\label{appendix:ctad_details}
CTAD maintains a consistent configuration across all experiments to ensure fair comparison. We set the number of reference samples $M = 20$, the number of K-means centroids $K = 5$ and the calibration weight $\lambda = 1.0$ as default values, which we find provide a good balance between calibration quality and computational efficiency (see sensitivity analysis in Section~\ref{sec:sensitivity}). Importantly, for each dataset, the computed calibration term $\{\Delta_{test}\}$ over the test set is identical across all base TAD models, demonstrating CTAD's true model-agnostic property---the calibration mechanism operates independently of the underlying detector architecture. Algorithm~\ref{algorithm_ctad} summarizes the complete CTAD workflow.

\begin{algorithm}[t]
\caption{CTAD Inference Workflow}
\label{algorithm_ctad}
\begin{algorithmic}[1]
\INPUT Training set $\mathcal{D}_{train}$, test set $\mathcal{D}_{test}$, trained detector $f(\cdot): \mathbf{x} \to [0,1]$
\State \textbf{Offline:} Apply K-means (Eq.~\ref{eq:kmeans}) to obtain $\mQ = \frac{1}{K}\sum_{j=1}^{K} \delta _{C_j}$
\State \textbf{Offline:} Sample reference set $\{\mathbf{x}_i\}_{i=1}^M \sim \mathcal{D}_{train}$ 
\For{each $\mathbf{x}_{test}$ in $\mathcal{D}_{test}$}
    \State Obtain base score: $s_{test} \gets f(\mathbf{x}_{test})$
    \State Extend distribution: $\mP \gets \frac{1}{M+1} (\sum_{i=1}^{M} \delta _{\mathbf{x}_i} + \delta_{\mathbf{x}_{test}})$ (Eq.~\ref{eq:extend_p})
    \State Compute calibration: $\Delta_{test} \gets \mathbf{OT}(\mP, \mQ)$ (Eq.~\ref{eq:ot_distance})
    \State Update score: $s_{test}^* \gets s_{test} + \Delta_{test}$ (Eq.~\ref{eq:calibration_formula})
\EndFor
\OUTPUT Calibrated scores $\{s_{test}^*\}$
\end{algorithmic}
\end{algorithm}

% \clearpage
\section{Additional Results}

\subsection{Full Results of Improvement over Base Detectors}
\label{appendix:full_improv}
We evaluate the effectiveness of CTAD by measuring its improvements over the original source detectors across 34 benchmark datasets. The full results are provided in Table~\ref{appendix:full_aucpr} and Table~\ref{appendix:full_aucroc}.

\subsection{Full Results of Comparison with Other Baseline Methods}
\label{appendix:full_comparison}
We further compare the performance of CTAD with conventional and advanced TAD baselines.
The full results are detailed in Tables~\ref{appendix:methodsPR} and~\ref{appendix:methodsROC}.

\subsection{Full Results of Experimental Validation of Theoretical Foundations}
\label{appendix:full_theory_evidence}
A key theoretical assumption underlying CTAD (formalized in Theorem 1) is that anomalies exhibit larger OT distances than normal samples: $\mathbb{E}_{\mathbf{x}_{test} \in \mathcal{A}}[\mathbf{OT}(\mP, \mQ)] > \mathbb{E}_{\mathbf{x}_{test} \in \mathcal{N}}[\mathbf{OT}(\mP, \mQ)]$. We empirically validate this foundational property across all datasets, providing crucial evidence that our theoretical framework holds in practice. Table~\ref{appendix:ot_difference} presents the results from all 34 datasets.

\begin{table*}
\centering
\footnotesize
\caption{The performance improvement achieved by CTAD over trained source detectors ($f$) on 34 datasets in terms of AUC-PR ($\uparrow$).}
\label{appendix:full_aucpr}
\begin{tabular}{c|cc|cc|cc|cc|cc|cc|cc}
\toprule
Dataset & OCSVM & +\textbf{CTAD} & PCA & +\textbf{CTAD} & IForest & +\textbf{CTAD} & ECOD & +\textbf{CTAD} & KNN & +\textbf{CTAD} & MCM & +\textbf{CTAD} & DRL & +\textbf{CTAD} \\
\midrule
Abalone & 0.8459 & 0.8022 & 0.8393 & \textbf{0.8586} & 0.8481 & 0.8458 & 0.6554 & \textbf{0.7651} & 0.8596 & \textbf{0.8846} & 0.7471 & \textbf{0.8831} & 0.8837 & \textbf{0.8852} \\
Amazon & 0.1050 & \textbf{0.1056} & 0.1072 & \textbf{0.1202} & 0.1091 & 0.1077 & 0.1040 & \textbf{0.1068} & 0.0904 & \textbf{0.1154} & 0.1083 & \textbf{0.1141} & 0.1258 & 0.1226 \\
Annthyroid & 0.1831 & \textbf{0.2546} & 0.5657 & 0.4177 & 0.6149 & 0.5212 & 0.4002 & 0.3691 & 0.3525 & \textbf{0.3775} & 0.3215 & \textbf{0.6003} & 0.6757 & 0.6757 \\
Arrhythmia & 0.5339 & \textbf{0.6178} & 0.5336 & \textbf{0.5719} & 0.5097 & \textbf{0.6384} & 0.4461 & \textbf{0.5889} & 0.6008 & \textbf{0.6258} & 0.6107 & \textbf{0.6138} & 0.6293 & \textbf{0.6625} \\
Breastw & 0.9934 & 0.9814 & 0.9934 & 0.9912 & 0.9449 & \textbf{0.9926} & 0.9522 & 0.9410 & 0.9712 & \textbf{0.9962} & 0.9952 & 0.9872 & 0.9964 & 0.9964 \\
Cardio & 0.8614 & 0.8601 & 0.8628 & 0.8539 & 0.7018 & \textbf{0.8065} & 0.3636 & \textbf{0.8395} & 0.7580 & \textbf{0.8085} & 0.8489 & \textbf{0.8531} & 0.8314 & \textbf{0.8639} \\
Cardiotocography & 0.6619 & 0.5860 & 0.6969 & 0.6861 & 0.6036 & \textbf{0.7424} & 0.6968 & \textbf{0.6987} & 0.6162 & \textbf{0.6895} & 0.6993 & \textbf{0.7652} & 0.7516 & \textbf{0.7831} \\
Comm.and.crime & 0.8371 & \textbf{0.8449} & 0.8892 & 0.8863 & 0.8940 & \textbf{0.8949} & 0.6854 & \textbf{0.7903} & 0.8510 & \textbf{0.8622} & 0.8549 & \textbf{0.8818} & 0.9160 & 0.9160 \\
Fault & 0.6062 & \textbf{0.6168} & 0.6035 & \textbf{0.6589} & 0.5948 & \textbf{0.6362} & 0.5171 & \textbf{0.5899} & 0.6028 & \textbf{0.6234} & 0.6022 & \textbf{0.6587} & 0.6579 & \textbf{0.6950} \\
Glass & 0.0896 & \textbf{0.1147} & 0.0896 & \textbf{0.1012} & 0.0952 & \textbf{0.1109} & 0.1113 & \textbf{0.1196} & 0.1099 & \textbf{0.1183} & 0.1905 & 0.1070 & 0.1526 & 0.1526 \\
Hepatitis & 0.2815 & \textbf{0.4500} & 0.5828 & \textbf{0.7394} & 0.4182 & 0.4159 & 0.4049 & \textbf{0.4651} & 0.2744 & \textbf{0.4317} & 0.3372 & \textbf{0.5929} & 0.6299 & \textbf{0.8005} \\
Imgseg & 0.7883 & \textbf{0.8622} & 0.7724 & \textbf{0.7887} & 0.7556 & \textbf{0.8159} & 0.7365 & \textbf{0.8092} & 0.8531 & \textbf{0.8643} & 0.8124 & \textbf{0.8655} & 0.9219 & 0.9219 \\
Ionosphere & 0.8969 & \textbf{0.9533} & 0.8969 & \textbf{0.9859} & 0.9768 & 0.9291 & 0.9713 & 0.9018 & 0.9297 & \textbf{0.9529} & 0.9802 & 0.9598 & 0.9897 & \textbf{0.9934} \\
Lympho & 0.8107 & \textbf{0.9306} & 1.0000 & \textbf{1.0000} & 0.9593 & \textbf{1.0000} & 0.8972 & \textbf{0.8972} & 0.9401 & \textbf{0.9484} & 0.4204 & \textbf{1.0000} & 1.0000 & \textbf{1.0000} \\
Mammography & 0.4178 & 0.3600 & 0.4165 & \textbf{0.5226} & 0.3334 & \textbf{0.5109} & 0.5380 & 0.4681 & 0.3810 & \textbf{0.3893} & 0.4755 & 0.3764 & 0.5482 & \textbf{0.5562} \\
Mnist & 0.1686 & \textbf{0.5410} & 0.6499 & 0.5957 & 0.5349 & \textbf{0.5899} & 0.3018 & \textbf{0.5544} & 0.7610 & \textbf{0.7722} & 0.7782 & \textbf{0.7821} & 0.9012 & 0.9012 \\
Musk & 0.0614 & \textbf{0.1168} & 1.0000 & \textbf{1.0000} & 0.5279 & \textbf{0.5766} & 0.9820 & 0.8465 & 0.9917 & \textbf{1.0000} & 0.6390 & \textbf{1.0000} & 1.0000 & \textbf{1.0000} \\
Optdigits & 0.0692 & \textbf{0.3890} & 0.0602 & \textbf{0.6546} & 0.1570 & \textbf{0.2053} & 0.0669 & \textbf{0.1719} & 0.8589 & \textbf{0.8672} & 0.8885 & \textbf{0.8929} & 0.6673 & \textbf{0.8140} \\
Parkinson & 0.8892 & 0.8675 & 0.9297 & 0.9147 & 0.9595 & 0.9311 & 0.8906 & 0.8855 & 0.7952 & \textbf{0.8495} & 0.7996 & \textbf{0.9379} & 0.9421 & 0.9421 \\
Pendigits & 0.5178 & \textbf{0.5348} & 0.3863 & \textbf{0.7676} & 0.5133 & \textbf{0.6376} & 0.4145 & 0.3821 & 0.9692 & \textbf{0.9792} & 0.8258 & \textbf{0.8804} & 0.9063 & \textbf{0.9600} \\
Pima & 0.7008 & 0.6209 & 0.7008 & \textbf{0.7380} & 0.6662 & \textbf{0.7433} & 0.5877 & \textbf{0.6601} & 0.7098 & \textbf{0.7181} & 0.7389 & 0.7387 & 0.7447 & \textbf{0.7488} \\
Satellite & 0.7778 & \textbf{0.7934} & 0.7778 & \textbf{0.8496} & 0.8583 & 0.8092 & 0.8334 & 0.8050 & 0.8515 & \textbf{0.8598} & 0.8532 & \textbf{0.8881} & 0.8518 & \textbf{0.8850} \\
Satimage-2 & 0.9192 & \textbf{0.9495} & 0.9192 & \textbf{0.9674} & 0.8846 & \textbf{0.9671} & 0.7775 & \textbf{0.9524} & 0.9555 & \textbf{0.9646} & 0.9850 & 0.8654 & 0.9659 & \textbf{0.9768} \\
Shuttle & 0.9488 & \textbf{0.9513} & 0.9627 & 0.7070 & 0.9172 & \textbf{0.9887} & 0.9815 & 0.8133 & 0.9370 & \textbf{0.9454} & 0.9479 & \textbf{0.9843} & 0.9766 & 0.9766 \\
Speech & 0.0279 & 0.0250 & 0.0277 & 0.0276 & 0.0353 & 0.0257 & 0.0287 & \textbf{0.0289} & 0.0197 & \textbf{0.0287} & 0.0380 & 0.0288 & 0.0533 & 0.0533 \\
Thyroid & 0.8134 & 0.5087 & 0.8134 & 0.6670 & 0.6055 & \textbf{0.7830} & 0.6807 & 0.6270 & 0.5903 & \textbf{0.5986} & 0.8417 & 0.7836 & 0.8745 & 0.8745 \\
Vertebral & 0.1517 & \textbf{0.1837} & 0.1381 & \textbf{0.2688} & 0.1342 & \textbf{0.1356} & 0.1917 & 0.1788 & 0.1239 & \textbf{0.1512} & 0.1949 & 0.1927 & 0.2796 & \textbf{0.3918} \\
Vowels & 0.2969 & \textbf{0.3226} & 0.1051 & \textbf{0.3921} & 0.0984 & \textbf{0.2403} & 0.1772 & \textbf{0.3405} & 0.3146 & \textbf{0.3872} & 0.0977 & \textbf{0.1746} & 0.4814 & \textbf{0.5312} \\
Wbc & 0.8391 & \textbf{0.8927} & 0.8391 & \textbf{0.9051} & 0.8573 & \textbf{0.8858} & 0.7217 & \textbf{0.7882} & 0.8022 & \textbf{0.8134} & 0.8887 & \textbf{0.8931} & 0.9704 & \textbf{1.0000} \\
WDBC & 0.4348 & \textbf{0.9573} & 0.9833 & \textbf{0.9909} & 0.9749 & \textbf{1.0000} & 0.7734 & \textbf{0.9698} & 0.9323 & \textbf{0.9573} & 0.8890 & \textbf{0.8934} & 1.0000 & \textbf{1.0000} \\
Wilt & 0.2254 & 0.1047 & 0.0641 & \textbf{0.0840} & 0.0848 & 0.0814 & 0.0768 & \textbf{0.0819} & 0.1496 & \textbf{0.1746} & 0.0759 & \textbf{0.1549} & 0.4178 & 0.4178 \\
Wine & 0.1424 & \textbf{0.6683} & 0.1325 & \textbf{0.2656} & 0.2458 & \textbf{0.5908} & 0.3578 & \textbf{0.6914} & 0.9917 & \textbf{1.0000} & 0.9335 & \textbf{0.9382} & 1.0000 & \textbf{1.0000} \\
WPBC & 0.3974 & 0.3962 & 0.3940 & \textbf{0.4150} & 0.3760 & \textbf{0.4030} & 0.3525 & \textbf{0.3784} & 0.3882 & \textbf{0.4132} & 0.4020 & \textbf{0.4422} & 0.4984 & \textbf{0.5109} \\
Yeast & 0.4803 & \textbf{0.4812} & 0.4678 & \textbf{0.5575} & 0.4654 & \textbf{0.4883} & 0.4943 & \textbf{0.5068} & 0.4737 & \textbf{0.5061} & 0.4631 & \textbf{0.5198} & 0.5433 & \textbf{0.5801} \\ \midrule
Average AUC-PR & 0.5228 & \textbf{0.5778} & 0.5942 & \textbf{0.6456} & 0.5664 & \textbf{0.6191} & 0.5344 & \textbf{0.5886} & 0.6414 & \textbf{0.6669} & 0.6260 & \textbf{0.6838} & 0.7290 & \textbf{0.7526} \\
\bottomrule
\end{tabular}
\end{table*}

\begin{table*}
\centering
\footnotesize
\caption{The performance improvement achieved by CTAD over trained source detectors ($f$) on 34 datasets in terms of AUC-ROC ($\uparrow$).}
\label{appendix:full_aucroc}
\begin{tabular}{c|cc|cc|cc|cc|cc|cc|cc}
\toprule
Dataset & OCSVM & +\textbf{CTAD} & PCA & +\textbf{CTAD} & IForest & +\textbf{CTAD} & ECOD & +\textbf{CTAD} & KNN & +\textbf{CTAD} & MCM & +\textbf{CTAD} & DRL & +\textbf{CTAD} \\
\midrule
Abalone & 0.7237 & 0.6580 & 0.7044 & \textbf{0.7366} & 0.7351 & \textbf{0.7413} & 0.4867 & \textbf{0.5809} & 0.7894 & \textbf{0.8144} & 0.5822 & \textbf{0.7913} & 0.8055 & 0.7984 \\
Amazon & 0.5418 & \textbf{0.5444} & 0.5490 & \textbf{0.5836} & 0.5593 & 0.5539 & 0.5379 & \textbf{0.5468} & 0.5768 & \textbf{0.6018} & 0.5615 & \textbf{0.5784} & 0.5806 & \textbf{0.5865} \\
Annthyroid & 0.5551 & \textbf{0.6362} & 0.8519 & 0.7184 & 0.9112 & 0.8728 & 0.7845 & 0.7549 & 0.6903 & \textbf{0.7153} & 0.6894 & \textbf{0.8916} & 0.9203 & 0.9203 \\
Arrhythmia & 0.7689 & \textbf{0.8225} & 0.7684 & \textbf{0.7927} & 0.7734 & \textbf{0.8184} & 0.7199 & \textbf{0.7996} & 0.7933 & \textbf{0.8183} & 0.8114 & \textbf{0.8155} & 0.7588 & \textbf{0.8048} \\
Breastw & 0.9938 & 0.9860 & 0.9938 & 0.9914 & 0.9719 & \textbf{0.9919} & 0.9649 & 0.9245 & 0.9714 & \textbf{0.9964} & 0.9955 & 0.9891 & 0.9965 & 0.9965 \\
Cardio & 0.9654 & 0.9644 & 0.9655 & 0.9559 & 0.9220 & \textbf{0.9354} & 0.6370 & \textbf{0.9640} & 0.8979 & \textbf{0.9250} & 0.9603 & \textbf{0.9651} & 0.9558 & \textbf{0.9658} \\
Cardiotocography & 0.7522 & 0.6482 & 0.7889 & \textbf{0.7907} & 0.7249 & \textbf{0.8409} & 0.7889 & \textbf{0.7956} & 0.6925 & \textbf{0.7450} & 0.8001 & \textbf{0.8091} & 0.8383 & \textbf{0.8486} \\
Comm.and.crime & 0.7047 & \textbf{0.7166} & 0.7871 & 0.7821 & 0.8044 & 0.8023 & 0.5170 & \textbf{0.6340} & 0.7269 & \textbf{0.7386} & 0.7361 & \textbf{0.7846} & 0.8458 & 0.8458 \\
Fault & 0.5682 & \textbf{0.5876} & 0.5587 & \textbf{0.6164} & 0.5609 & \textbf{0.6240} & 0.5037 & \textbf{0.5448} & 0.5795 & \textbf{0.6018} & 0.6062 & \textbf{0.6386} & 0.6204 & \textbf{0.6515} \\
Glass & 0.5480 & \textbf{0.5987} & 0.5480 & \textbf{0.5879} & 0.5771 & \textbf{0.5879} & 0.6235 & 0.5556 & 0.6141 & \textbf{0.6224} & 0.7225 & 0.6117 & 0.6828 & 0.6828 \\
Hepatitis & 0.4955 & \textbf{0.6425} & 0.8122 & \textbf{0.8167} & 0.7255 & 0.7081 & 0.6991 & 0.6810 & 0.4682 & \textbf{0.5769} & 0.5289 & \textbf{0.7941} & 0.8100 & \textbf{0.8281} \\
Imgseg & 0.7414 & \textbf{0.7938} & 0.6735 & \textbf{0.7484} & 0.6859 & \textbf{0.7175} & 0.6242 & \textbf{0.7349} & 0.8723 & \textbf{0.8809} & 0.8134 & \textbf{0.8242} & 0.9037 & 0.9037 \\
Ionosphere & 0.8765 & \textbf{0.9370} & 0.8765 & \textbf{0.9822} & 0.9683 & 0.9103 & 0.9569 & 0.8657 & 0.9167 & \textbf{0.9331} & 0.9726 & 0.9687 & 0.9841 & \textbf{0.9921} \\
Lympho & 0.9812 & \textbf{0.9930} & 1.0000 & \textbf{1.0000} & 0.9945 & \textbf{1.0000} & 0.9906 & \textbf{0.9906} & 0.9870 & \textbf{0.9953} & 0.9257 & \textbf{1.0000} & 1.0000 & \textbf{1.0000} \\
Mammography & 0.9003 & 0.8374 & 0.8993 & \textbf{0.9087} & 0.8220 & \textbf{0.8707} & 0.8251 & \textbf{0.8868} & 0.8640 & \textbf{0.8724} & 0.9053 & 0.8111 & 0.9122 & \textbf{0.9129} \\
Mnist & 0.5000 & \textbf{0.8348} & 0.9022 & 0.8639 & 0.8623 & \textbf{0.8833} & 0.7487 & \textbf{0.8738} & 0.9265 & \textbf{0.9370} & 0.9284 & \textbf{0.9330} & 0.9736 & 0.9736 \\
Musk & 0.5000 & \textbf{0.7738} & 1.0000 & \textbf{1.0000} & 0.9521 & \textbf{0.9704} & 0.9987 & 0.9876 & 0.9917 & \textbf{1.0000} & 0.9752 & \textbf{1.0000} & 1.0000 & \textbf{1.0000} \\
Optdigits & 0.6338 & \textbf{0.9352} & 0.5817 & \textbf{0.9758} & 0.8239 & \textbf{0.8785} & 0.6145 & \textbf{0.8458} & 0.9862 & \textbf{0.9945} & 0.9947 & \textbf{0.9997} & 0.9717 & \textbf{0.9887} \\
Parkinson & 0.6043 & \textbf{0.6162} & 0.6927 & 0.6837 & 0.7680 & 0.7163 & 0.5150 & \textbf{0.6137} & 0.4365 & \textbf{0.5275} & 0.3661 & \textbf{0.7341} & 0.7293 & 0.7293 \\
Pendigits & 0.9636 & \textbf{0.9703} & 0.9437 & \textbf{0.9921} & 0.9666 & \textbf{0.9760} & 0.9295 & \textbf{0.9471} & 0.9906 & \textbf{0.9988} & 0.9919 & \textbf{0.9960} & 0.9944 & \textbf{0.9990} \\
Pima & 0.7133 & 0.6415 & 0.7133 & \textbf{0.7211} & 0.6737 & \textbf{0.7367} & 0.5834 & \textbf{0.6242} & 0.6723 & \textbf{0.6807} & 0.7639 & 0.7511 & 0.7650 & 0.7198 \\
Satellite & 0.6663 & \textbf{0.7516} & 0.6663 & \textbf{0.7838} & 0.8026 & 0.7653 & 0.7884 & 0.7192 & 0.8139 & \textbf{0.8223} & 0.7962 & \textbf{0.8704} & 0.8018 & \textbf{0.8588} \\
Satimage-2 & 0.9817 & \textbf{0.9940} & 0.9817 & \textbf{0.9970} & 0.9938 & \textbf{0.9945} & 0.9650 & \textbf{0.9942} & 0.9891 & \textbf{0.9973} & 0.9992 & 0.9950 & 0.9956 & \textbf{0.9983} \\
Shuttle & 0.9969 & 0.9955 & 0.9936 & 0.8926 & 0.9961 & \textbf{0.9969} & 0.9978 & 0.9427 & 0.9893 & \textbf{0.9977} & 0.9975 & \textbf{0.9993} & 0.9989 & \textbf{0.9989} \\
Speech & 0.3673 & 0.3562 & 0.3638 & 0.3637 & 0.3812 & 0.3595 & 0.3596 & \textbf{0.3630} & 0.3561 & \textbf{0.3651} & 0.4409 & 0.3709 & 0.6073 & 0.6073 \\
Thyroid & 0.9855 & 0.9246 & 0.9855 & 0.9605 & 0.9271 & \textbf{0.9881} & 0.8827 & \textbf{0.9714} & 0.9525 & \textbf{0.9608} & 0.9804 & 0.9778 & 0.9933 & 0.9933 \\
Vertebral & 0.2654 & \textbf{0.4089} & 0.1746 & \textbf{0.5092} & 0.1444 & \textbf{0.1556} & 0.4124 & 0.3854 & 0.1171 & \textbf{0.2638} & 0.3767 & 0.3463 & 0.6156 & \textbf{0.6241} \\
Vowels & 0.7557 & \textbf{0.8127} & 0.5229 & \textbf{0.8236} & 0.5902 & \textbf{0.7639} & 0.6147 & \textbf{0.8164} & 0.8174 & \textbf{0.8714} & 0.6529 & \textbf{0.8340} & 0.8654 & \textbf{0.8942} \\
Wbc & 0.9667 & \textbf{0.9848} & 0.9667 & \textbf{0.9843} & 0.9715 & \textbf{0.9806} & 0.8747 & \textbf{0.9423} & 0.9536 & \textbf{0.9606} & 0.9814 & \textbf{0.9863} & 0.9960 & \textbf{1.0000} \\
WDBC & 0.9637 & \textbf{0.9978} & 0.9989 & \textbf{0.9994} & 0.9983 & \textbf{1.0000} & 0.9793 & \textbf{0.9983} & 0.9728 & \textbf{0.9978} & 0.9419 & \textbf{0.9466} & 1.0000 & \textbf{1.0000} \\
Wilt & 0.7923 & 0.5050 & 0.2607 & \textbf{0.4505} & 0.4595 & 0.4308 & 0.3748 & \textbf{0.4179} & 0.7295 & \textbf{0.7545} & 0.3741 & \textbf{0.7091} & 0.9278 & 0.9278 \\
Wine & 0.4850 & \textbf{0.8400} & 0.4467 & \textbf{0.5217} & 0.6571 & \textbf{0.7283} & 0.7433 & \textbf{0.8550} & 0.9917 & \textbf{1.0000} & 0.9538 & \textbf{0.9586} & 1.0000 & \textbf{1.0000} \\
WPBC & 0.4700 & \textbf{0.5210} & 0.4686 & \textbf{0.5482} & 0.4975 & \textbf{0.5344} & 0.4706 & \textbf{0.4908} & 0.4854 & \textbf{0.5104} & 0.5105 & \textbf{0.5412} & 0.5985 & \textbf{0.6408} \\
Yeast & 0.4483 & \textbf{0.4513} & 0.4324 & \textbf{0.5209} & 0.4095 & \textbf{0.4652} & 0.4464 & \textbf{0.4784} & 0.4366 & \textbf{0.4919} & 0.4259 & \textbf{0.4959} & 0.5134 & \textbf{0.5520} \\ \midrule
Average AUC-ROC & 0.7111 & \textbf{0.7553} & 0.7316 & \textbf{0.7825} & 0.7533 & \textbf{0.7735} & 0.7047 & \textbf{0.7508} & 0.7661 & \textbf{0.7932} & 0.7665 & \textbf{0.8152} & 0.8518 & \textbf{0.8601} \\
\bottomrule
\end{tabular}
\end{table*}

\clearpage
% \begin{sidewaystable}
\begin{table*}
\tiny
\centering
\caption{Comparison of AUC-PR ($\uparrow$) results between other baseline methods and CTAD.}
\label{appendix:methodsPR}
\centering
\resizebox{\linewidth}{!}{
\begin{tabular}{cccccccccccccc|ccc}
\toprule
                 & OCSVM   & KNN     & PCA     & IForest & ECOD    & DeepSVDD & AutoEncoder & GOAD    & NeuTraLAD & ICL     & DTE     & MCM     & DRL        & CTAD (KNN)& CTAD (MCM)& CTAD (DRL) \\ \midrule
Abalone          & 0.8459  & 0.8596  & 0.8393  & 0.8481  & 0.6554  & 0.7173   & 0.8295      & 0.8233  & 0.7768    & 0.7654  & 0.4798  & 0.7471  & 0.8837     & 0.8846    & 0.8831    & 0.8852    \\
Amazon           & 0.1050  & 0.0904  & 0.1072  & 0.1091  & 0.1040  & 0.1003   & 0.1170      & 0.1086  & 0.1007    & 0.1077  & 0.1115  & 0.1083  & 0.1258     & 0.1154    & 0.1141    & 0.1226    \\
Annthyroid       & 0.1831  & 0.3525  & 0.5657  & 0.6149  & 0.4002  & 0.3235   & 0.5291      & 0.2132  & 0.5057    & 0.4114  & 0.6288  & 0.3215  & 0.6757     & 0.3775    & 0.6003    & 0.6757    \\
Arrhythmia       & 0.5339  & 0.6008  & 0.5336  & 0.5097  & 0.4461  & 0.6036   & 0.3029      & 0.4091  & 0.6237    & 0.6155  & 0.4912  & 0.6107  & 0.6293     & 0.6258    & 0.6138    & 0.6625    \\
Breastw          & 0.9934  & 0.9712  & 0.9934  & 0.9449  & 0.9522  & 0.9924   & 0.9896      & 0.8335  & 0.9117    & 0.9656  & 0.8825  & 0.9952  & 0.9964     & 0.9962    & 0.9872    & 0.9964    \\
Cardio           & 0.8614  & 0.7580  & 0.8628  & 0.7018  & 0.3636  & 0.7880   & 0.4778      & 0.6225  & 0.4535    & 0.8037  & 0.6929  & 0.8489  & 0.8314     & 0.8085    & 0.8531    & 0.8639    \\
Cardiotocography & 0.6619  & 0.6162  & 0.6969  & 0.6036  & 0.6968  & 0.4602   & 0.4705      & 0.4647  & 0.6431    & 0.5955  & 0.5334  & 0.6993  & 0.7516     & 0.6895    & 0.7652    & 0.7831    \\
Comm.and.crime   & 0.8371  & 0.8510  & 0.8892  & 0.8940  & 0.6854  & 0.8239   & 0.8678      & 0.9330  & 0.9202    & 0.8962  & 0.7901  & 0.8549  & 0.9160     & 0.8622    & 0.8818    & 0.9160    \\
Fault            & 0.6062  & 0.6028  & 0.6035  & 0.5948  & 0.5171  & 0.5630   & 0.6501      & 0.5872  & 0.6083    & 0.5937  & 0.6393  & 0.6022  & 0.6579     & 0.6234    & 0.6587    & 0.6950    \\
Glass            & 0.0896  & 0.1099  & 0.0896  & 0.0952  & 0.1113  & 0.0912   & 0.1079      & 0.0948  & 0.1491    & 0.2573  & 0.2151  & 0.1905  & 0.1526     & 0.1183    & 0.1070    & 0.1526    \\
Hepatitis        & 0.2815  & 0.2744  & 0.5828  & 0.4182  & 0.4049  & 0.4264   & 0.4248      & 0.3393  & 0.3690    & 0.3357  & 0.6582  & 0.3372  & 0.6299     & 0.4317    & 0.5929    & 0.8005    \\
Imgseg           & 0.7883  & 0.8531  & 0.7724  & 0.7556  & 0.7365  & 0.6699   & 0.8391      & 0.7114  & 0.8747    & 0.8916  & 0.6846  & 0.8124  & 0.9219     & 0.8643    & 0.8655    & 0.9219    \\
Ionosphere       & 0.8969  & 0.9297  & 0.8969  & 0.9768  & 0.9713  & 0.8670   & 0.7328      & 0.9280  & 0.9355    & 0.9777  & 0.9683  & 0.9802  & 0.9897     & 0.9529    & 0.9598    & 0.9934    \\
Lympho           & 0.8107  & 0.9401  & 1.0000  & 0.9593  & 0.8972  & 0.9749   & 0.2709      & 0.7656  & 0.6460    & 0.6091  & 0.8677  & 0.4204  & 1.0000     & 0.9484    & 1.0000    & 1.0000    \\
Mammography      & 0.4178  & 0.3810  & 0.4165  & 0.3334  & 0.5380  & 0.4190   & 0.2530      & 0.2426  & 0.1326    & 0.1894  & 0.3985  & 0.4755  & 0.5482     & 0.3893    & 0.3764    & 0.5562    \\
Mnist            & 0.1686  & 0.7610  & 0.6499  & 0.5349  & 0.3018  & 0.4158   & 0.2790      & 0.7182  & 0.9021    & 0.8915  & 0.5226  & 0.7782  & 0.9012     & 0.7722    & 0.7821    & 0.9012    \\
Musk             & 0.0614  & 0.9917  & 1.0000  & 0.5279  & 0.9820  & 1.0000   & 1.0000      & 0.5372  & 1.0000    & 1.0000  & 1.0000  & 0.6390  & 1.0000     & 1.0000    & 1.0000    & 1.0000    \\
Optdigits        & 0.0692  & 0.8589  & 0.0602  & 0.1570  & 0.0669  & 0.1159   & 0.1418      & 0.0633  & 0.1709    & 0.1696  & 0.1534  & 0.8885  & 0.6673     & 0.8672    & 0.8929    & 0.8140    \\
Parkinson        & 0.8892  & 0.7952  & 0.9297  & 0.9595  & 0.8906  & 0.9192   & 0.9243      & 0.8413  & 0.8172    & 0.8568  & 0.7304  & 0.7996  & 0.9421     & 0.8495    & 0.9379    & 0.9421    \\
Pendigits        & 0.5178  & 0.9692  & 0.3863  & 0.5133  & 0.4145  & 0.0616   & 0.8904      & 0.0259  & 0.6930    & 0.4039  & 0.4844  & 0.8258  & 0.9063     & 0.9792    & 0.8804    & 0.9600    \\
Pima             & 0.7008  & 0.7098  & 0.7008  & 0.6662  & 0.5877  & 0.7165   & 0.7174      & 0.5027  & 0.6168    & 0.6965  & 0.6798  & 0.7389  & 0.7447     & 0.7181    & 0.7387    & 0.7488    \\
Satellite        & 0.7778  & 0.8515  & 0.7778  & 0.8583  & 0.8334  & 0.8217   & 0.8218      & 0.7786  & 0.8588    & 0.8799  & 0.8479  & 0.8532  & 0.8518     & 0.8598    & 0.8881    & 0.8850    \\
Satimage-2       & 0.9192  & 0.9555  & 0.9192  & 0.8846  & 0.7775  & 0.9427   & 0.9688      & 0.9726  & 0.9684    & 0.8124  & 0.6821  & 0.9850  & 0.9659     & 0.9646    & 0.8654    & 0.9768    \\
Shuttle          & 0.9488  & 0.9370  & 0.9627  & 0.9172  & 0.9815  & 0.9818   & 0.9316      & 0.9545  & 0.9971    & 0.9811  & 0.9403  & 0.9479  & 0.9766     & 0.9454    & 0.9843    & 0.9766    \\
Speech           & 0.0279  & 0.0197  & 0.0277  & 0.0353  & 0.0287  & 0.0400   & 0.0270      & 0.0424  & 0.0386    & 0.0335  & 0.0285  & 0.0380  & 0.0533     & 0.0287    & 0.0288    & 0.0533    \\
Thyroid          & 0.8134  & 0.5903  & 0.8134  & 0.6055  & 0.6807  & 0.7282   & 0.8096      & 0.6931  & 0.7435    & 0.6575  & 0.8167  & 0.8417  & 0.8745     & 0.5986    & 0.7836    & 0.8745    \\
Vertebral        & 0.1517  & 0.1239  & 0.1381  & 0.1342  & 0.1917  & 0.1590   & 0.1476      & 0.1451  & 0.1658    & 0.1598  & 0.2514  & 0.1949  & 0.2796     & 0.1512    & 0.1927    & 0.3918    \\
Vowels           & 0.2969  & 0.3146  & 0.1051  & 0.0984  & 0.1772  & 0.1717   & 0.3475      & 0.2274  & 0.1224    & 0.1574  & 0.3810  & 0.0977  & 0.4814     & 0.3872    & 0.1746    & 0.5312    \\
Wbc              & 0.8391  & 0.8022  & 0.8391  & 0.8573  & 0.7217  & 0.8340   & 0.8578      & 0.3362  & 0.6051    & 0.7218  & 0.3997  & 0.8887  & 0.9704     & 0.8134    & 0.8931    & 1.0000    \\
WDBC             & 0.4348  & 0.9323  & 0.9833  & 0.9749  & 0.7734  & 0.9184   & 0.9591      & 0.9276  & 0.6372    & 0.8821  & 0.6882  & 0.8890  & 1.0000     & 0.9573    & 0.8934    & 1.0000    \\
Wilt             & 0.2254  & 0.1496  & 0.0641  & 0.0848  & 0.0768  & 0.0705   & 0.0867      & 0.1061  & 0.2440    & 0.1382  & 0.2541  & 0.0759  & 0.4178     & 0.1746    & 0.1549    & 0.4178    \\
Wine             & 0.1424  & 0.9917  & 0.1325  & 0.2458  & 0.3578  & 0.1476   & 0.1605      & 0.1619  & 0.9078    & 0.5659  & 0.9985  & 0.9335  & 1.0000     & 1.0000    & 0.9382    & 1.0000    \\
WPBC             & 0.3974  & 0.3882  & 0.3940  & 0.3760  & 0.3525  & 0.3903   & 0.3848      & 0.3738  & 0.3631    & 0.4238  & 0.4935  & 0.4020  & 0.4984     & 0.4132    & 0.4422    & 0.5109    \\
Yeast            & 0.4803  & 0.4737  & 0.4678  & 0.4654  & 0.4943  & 0.4953   & 0.4833      & 0.5877  & 0.5458    & 0.5097  & 0.4974  & 0.4631  & 0.5433     & 0.5061    & 0.5198    & 0.5801    \\ \midrule
Average          & 0.5228  & 0.6414  & 0.5942  & 0.5664  & 0.5344  & 0.5515   & 0.5530      & 0.5021  & 0.5897    & 0.5870  & 0.5851  & 0.6260  & 0.7290     & 0.6669    & 0.6838    & 0.7526    \\
Rank             & 15.6471 & 14.3824 & 13.5588 & 14.8235 & 16.5294 & 15.3971  & 13.8676     & 16.8235 & 12.6324   & 12.7794 & 13.4265 & 11.5588 & 3.8971     & 9.3529    & 7.9118    & 2.5441    \\ 
\bottomrule
% Win              & 0       & 0       & 2       & 0       & 0       & 1        & 1           & 2       & 3         & 2       & 1       & 0       & 9          & 3         & 3         & 18       
\end{tabular}}
\end{table*}
% \end{sidewaystable}

% \clearpage
% \begin{sidewaystable}
\begin{table*}
\tiny
\centering
\caption{Comparison of AUC-ROC ($\uparrow$) results between other baseline methods and CTAD.}
\label{appendix:methodsROC}
\centering
\resizebox{\linewidth}{!}{
\begin{tabular}{cccccccccccccc|ccc}
\toprule
                 & OCSVM   & KNN     & PCA     & IForest & ECOD    & DeepSVDD & AutoEncoder & GOAD    & NeuTraLAD & ICL     & DTE     & MCM     & DRL        & CTAD (KNN)& CTAD (MCM)& CTAD (DRL) \\ \midrule
Abalone          & 0.7237  & 0.7894  & 0.7044  & 0.7351  & 0.4867  & 0.5468   & 0.7019      & 0.7407  & 0.6992    & 0.6208  & 0.4624  & 0.5822  & 0.8055     & 0.8144    & 0.7913    & 0.7984    \\
Amazon           & 0.5418  & 0.5768  & 0.5490  & 0.5593  & 0.5379  & 0.5095   & 0.6004      & 0.5584  & 0.5187    & 0.5596  & 0.5671  & 0.5615  & 0.5806     & 0.6018    & 0.5784    & 0.5865    \\
Annthyroid       & 0.5551  & 0.6903  & 0.8519  & 0.9112  & 0.7845  & 0.5678   & 0.7295      & 0.6008  & 0.8196    & 0.6997  & 0.9090  & 0.6894  & 0.9203     & 0.7153    & 0.8916    & 0.9203    \\
Arrhythmia       & 0.7689  & 0.7933  & 0.7684  & 0.7734  & 0.7199  & 0.7941   & 0.5682      & 0.5810  & 0.7900    & 0.8145  & 0.5912  & 0.8114  & 0.7588     & 0.8183    & 0.8155    & 0.8048    \\
Breastw          & 0.9938  & 0.9714  & 0.9938  & 0.9719  & 0.9649  & 0.9925   & 0.9909      & 0.7327  & 0.9458    & 0.9725  & 0.9278  & 0.9955  & 0.9965     & 0.9964    & 0.9891    & 0.9965    \\
Cardio           & 0.9654  & 0.8979  & 0.9655  & 0.9220  & 0.6370  & 0.9313   & 0.5670      & 0.6812  & 0.7349    & 0.9514  & 0.8726  & 0.9603  & 0.9558     & 0.9250    & 0.9651    & 0.9658    \\
Cardiotocography & 0.7522  & 0.6925  & 0.7889  & 0.7249  & 0.7889  & 0.5149   & 0.5227      & 0.4428  & 0.7177    & 0.6478  & 0.6013  & 0.8001  & 0.8383     & 0.7450    & 0.8091    & 0.8486    \\
Comm.and.crime   & 0.7047  & 0.7269  & 0.7871  & 0.8044  & 0.5170  & 0.6941   & 0.7449      & 0.8785  & 0.8517    & 0.8213  & 0.6134  & 0.7361  & 0.8458     & 0.7386    & 0.7846    & 0.8458    \\
Fault            & 0.5682  & 0.5795  & 0.5587  & 0.5609  & 0.5037  & 0.5342   & 0.6073      & 0.5742  & 0.5753    & 0.5777  & 0.5864  & 0.6062  & 0.6204     & 0.6018    & 0.6386    & 0.6515    \\
Glass            & 0.5480  & 0.6141  & 0.5480  & 0.5771  & 0.6235  & 0.5566   & 0.6088      & 0.5741  & 0.6312    & 0.8350  & 0.6964  & 0.7225  & 0.6828     & 0.6224    & 0.6117    & 0.6828    \\
Hepatitis        & 0.4955  & 0.4682  & 0.8122  & 0.7255  & 0.6991  & 0.6621   & 0.5977      & 0.5724  & 0.6584    & 0.5618  & 0.8022  & 0.5289  & 0.8100     & 0.5769    & 0.7941    & 0.8281    \\
Imgseg           & 0.7414  & 0.8723  & 0.6735  & 0.6859  & 0.6242  & 0.5348   & 0.7991      & 0.6824  & 0.8726    & 0.8717  & 0.7252  & 0.8134  & 0.9037     & 0.8809    & 0.8242    & 0.9037    \\
Ionosphere       & 0.8765  & 0.9167  & 0.8765  & 0.9683  & 0.9569  & 0.8552   & 0.6231      & 0.8993  & 0.9453    & 0.9710  & 0.9542  & 0.9726  & 0.9841     & 0.9331    & 0.9687    & 0.9921    \\
Lympho           & 0.9812  & 0.9870  & 1.0000  & 0.9945  & 0.9906  & 0.9977   & 0.7856      & 0.9757  & 0.9648    & 0.9546  & 0.9899  & 0.9257  & 1.0000     & 0.9953    & 1.0000    & 1.0000    \\
Mammography      & 0.9003  & 0.8640  & 0.8993  & 0.8220  & 0.8251  & 0.8879   & 0.8472      & 0.7348  & 0.7449    & 0.6548  & 0.8862  & 0.9053  & 0.9122     & 0.8724    & 0.8111    & 0.9129    \\
Mnist            & 0.5000  & 0.9265  & 0.9022  & 0.8623  & 0.7487  & 0.6371   & 0.5151      & 0.9201  & 0.9737    & 0.9646  & 0.8743  & 0.9284  & 0.9736     & 0.9370    & 0.9330    & 0.9736    \\
Musk             & 0.5000  & 0.9917  & 1.0000  & 0.9521  & 0.9987  & 1.0000   & 1.0000      & 0.9543  & 1.0000    & 1.0000  & 1.0000  & 0.9752  & 1.0000     & 1.0000    & 1.0000    & 1.0000    \\
Optdigits        & 0.6338  & 0.9862  & 0.5817  & 0.8239  & 0.6145  & 0.7603   & 0.6694      & 0.5972  & 0.8471    & 0.7870  & 0.8238  & 0.9947  & 0.9717     & 0.9945    & 0.9997    & 0.9887    \\
Parkinson        & 0.6043  & 0.4365  & 0.6927  & 0.7680  & 0.5150  & 0.6601   & 0.6943      & 0.4688  & 0.4205    & 0.4747  & 0.4672  & 0.3661  & 0.7293     & 0.5275    & 0.7341    & 0.7293    \\
Pendigits        & 0.9636  & 0.9906  & 0.9437  & 0.9666  & 0.9295  & 0.4563   & 0.9937      & 0.2141  & 0.9859    & 0.9142  & 0.9761  & 0.9919  & 0.9944     & 0.9988    & 0.9960    & 0.9990    \\
Pima             & 0.7133  & 0.6723  & 0.7133  & 0.6737  & 0.5834  & 0.7348   & 0.7163      & 0.4338  & 0.6170    & 0.6727  & 0.6788  & 0.7639  & 0.7650     & 0.6807    & 0.7511    & 0.7198    \\
Satellite        & 0.6663  & 0.8139  & 0.6663  & 0.8026  & 0.7884  & 0.7659   & 0.7233      & 0.7374  & 0.8080    & 0.8549  & 0.7661  & 0.7962  & 0.8018     & 0.8223    & 0.8704    & 0.8588    \\
Satimage-2       & 0.9817  & 0.9891  & 0.9817  & 0.9938  & 0.9650  & 0.9881   & 0.9979      & 0.9929  & 0.9979    & 0.9792  & 0.9967  & 0.9992  & 0.9956     & 0.9973    & 0.9950    & 0.9983    \\
Shuttle          & 0.9969  & 0.9893  & 0.9936  & 0.9961  & 0.9978  & 0.9952   & 0.9944      & 0.9897  & 0.9994    & 0.9935  & 0.9993  & 0.9975  & 0.9989     & 0.9977    & 0.9993    & 0.9989    \\
Speech           & 0.3673  & 0.3561  & 0.3638  & 0.3812  & 0.3596  & 0.5071   & 0.3633      & 0.5065  & 0.4809    & 0.4883  & 0.3817  & 0.4409  & 0.6073     & 0.3651    & 0.3709    & 0.6073    \\
Thyroid          & 0.9855  & 0.9525  & 0.9855  & 0.9271  & 0.8827  & 0.9887   & 0.9780      & 0.8523  & 0.9701    & 0.9518  & 0.9863  & 0.9804  & 0.9933     & 0.9608    & 0.9778    & 0.9933    \\
Vertebral        & 0.2654  & 0.1171  & 0.1746  & 0.1444  & 0.4124  & 0.2706   & 0.2375      & 0.2262  & 0.3121    & 0.2821  & 0.5430  & 0.3767  & 0.6156     & 0.2638    & 0.3463    & 0.6241    \\
Vowels           & 0.7557  & 0.8174  & 0.5229  & 0.5902  & 0.6147  & 0.5734   & 0.7905      & 0.7247  & 0.7239    & 0.7020  & 0.8142  & 0.6529  & 0.8654     & 0.8714    & 0.8340    & 0.8942    \\
Wbc              & 0.9667  & 0.9536  & 0.9667  & 0.9715  & 0.8747  & 0.9633   & 0.9737      & 0.4806  & 0.9133    & 0.9080  & 0.8054  & 0.9814  & 0.9960     & 0.9606    & 0.9863    & 1.0000    \\
WDBC             & 0.9637  & 0.9728  & 0.9989  & 0.9983  & 0.9793  & 0.9879   & 0.9976      & 0.9950  & 0.9568    & 0.9898  & 0.9852  & 0.9419  & 1.0000     & 0.9978    & 0.9466    & 1.0000    \\
Wilt             & 0.7923  & 0.7295  & 0.2607  & 0.4595  & 0.3748  & 0.3420   & 0.4633      & 0.5589  & 0.8328    & 0.5332  & 0.6291  & 0.3741  & 0.9278     & 0.7545    & 0.7091    & 0.9278    \\
Wine             & 0.4850  & 0.9917  & 0.4467  & 0.6571  & 0.7433  & 0.5067   & 0.5356      & 0.5366  & 0.9753    & 0.9150  & 0.9944  & 0.9538  & 1.0000     & 1.0000    & 0.9586    & 1.0000    \\
WPBC             & 0.4700  & 0.4854  & 0.4686  & 0.4975  & 0.4706  & 0.4956   & 0.4942      & 0.4591  & 0.4709    & 0.5482  & 0.5916  & 0.5105  & 0.5985     & 0.5104    & 0.5412    & 0.6408    \\
Yeast            & 0.4483  & 0.4366  & 0.4324  & 0.4095  & 0.4464  & 0.4790   & 0.4503      & 0.6000  & 0.5488    & 0.5076  & 0.4458  & 0.4259  & 0.5134     & 0.4919    & 0.4959    & 0.5520    \\ \midrule
Average          & 0.7111  & 0.7661  & 0.7316  & 0.7533  & 0.7047  & 0.6968   & 0.7024      & 0.6611  & 0.7737    & 0.7641  & 0.7631  & 0.7665  & 0.8518     & 0.7932    & 0.8152    & 0.8601    \\
Rank             & 15.9118 & 14.3529 & 14.4853 & 14.0000 & 17.4265 & 15.7353  & 14.5000     & 17.6765 & 12.3088   & 13.1324 & 12.6618 & 11.4118 & 4.2941     & 9.2059    & 7.5294    & 2.6765    \\
% Win              & 0       & 0       & 2       & 0       & 0       & 1        & 1           & 2       & 3         & 2       & 1       & 0       & 11         & 2         & 4         & 19       
\bottomrule
\end{tabular}}
\end{table*}
% \end{sidewaystable}

\begin{table*}
\footnotesize
\centering
\caption{
The difference between $\mathbb{E}_{\mathbf{x}_{test} \in \mathcal{N}}[\mathbf{OT}(\mP, \mQ)]$ and $\mathbb{E}_{\mathbf{x}_{test} \in \mathcal{A}}[\mathbf{OT}(\mP, \mQ)]$ across different datasets.
The results are calculated based on $\mathcal{D}_{test}$.
``Increase (\%)'' indicates the relative improvement from $\mathbb{E}_{\mathbf{x}_{test} \in \mathcal{N}}[\mathbf{OT}(\mP, \mQ)]$ to $\mathbb{E}_{\mathbf{x}_{test} \in \mathcal{A}}[\mathbf{OT}(\mP, \mQ)]$, i.e., $\frac{(\mathbb{E}_{\mathbf{x}_{test} \in \mathcal{A}}[\mathbf{OT}(\mP, \mQ)] - \mathbb{E}_{\mathbf{x}_{test} \in \mathcal{N}}[\mathbf{OT}(\mP, \mQ)])}{\mathbb{E}_{\mathbf{x}_{test} \in \mathcal{N}}[\mathbf{OT}(\mP, \mQ)]}$. 
Anomalies exhibit substantially higher OT distances on 31/34 datasets (91.2\%).
}
\label{appendix:ot_difference}
\setlength{\tabcolsep}{2.3mm}{
\begin{tabular}{cccccccccc}
\toprule
  & abalone   & amazon    & annthyroid & arrhythmia & breastw    & cardio      & Cardiotocography & comm.and.crime & fault      \\ \midrule
$\mathbb{E}_{\mathbf{x}^* \in \mathcal{N}}[\mathbf{OT}(\mP, \mQ)]$ & 0.0179    & 0.0338    & 0.0007     & 1.4669     & 0.0577     & 0.044       & 2.1373           & 0.0713         & 0.2319     \\
$\mathbb{E}_{\mathbf{x}^* \in \mathcal{A}}[\mathbf{OT}(\mP, \mQ)]$ & 0.0512    & 0.035     & 0.0011     & 2.3242     & 0.087      & 0.0895      & 4.0559           & 0.1046         & 0.375      \\
Increase (\%) & 185.3388  & 3.3587    & 50.4653    & 58.4472    & 50.7674    & 103.3323    & 89.7682          & 46.7356        & 61.7357    \\ \bottomrule \toprule
  & glass     & Hepatitis & imgseg     & ionosphere & lympho     & mammography & mnist            & musk           & optdigits  \\ \midrule
$\mathbb{E}_{\mathbf{x}^* \in \mathcal{N}}[\mathbf{OT}(\mP, \mQ)]$  & 0.0279    & 1.329     & 0.0418     & 0.0146     & 0.0206     & 0.0213      & 5.2837           & 10.2871        & 0.2858     \\
$\mathbb{E}_{\mathbf{x}^* \in \mathcal{A}}[\mathbf{OT}(\mP, \mQ)]$  & 0.0319    & 1.6077    & 0.1331     & 0.0389     & 0.0533     & 0.0571      & 6.7088           & 15.6739        & 0.354      \\ 
Increase (\%)  & 14.2997   & 20.9713   & 218.7097   & 165.8724   & 158.8689   & 168.8426    & 26.9707          & 52.3637        & 23.8584    \\ \bottomrule \toprule
  & Parkinson & pendigits & pima       & satellite  & satimage-2 & shuttle     & speech           & thyroid        & vertebral  \\ \midrule
$\mathbb{E}_{\mathbf{x}^* \in \mathcal{N}}[\mathbf{OT}(\mP, \mQ)]$  & 0.8832    & 0.008     & 0.9061     & 0.5816     & 0.5986     & 50.7923     & 0.1961           & 0.0013         & 0.4131     \\
$\mathbb{E}_{\mathbf{x}^* \in \mathcal{A}}[\mathbf{OT}(\mP, \mQ)]$  & 0.979     & 0.0178    & 1.964      & 1.2497     & 2.1898     & 111.8185    & 0.18             & 0.0034         & 0.276      \\
Increase (\%)  & 10.8526   & 123.3686  & 116.7628   & 114.8635   & 265.805    & 120.1483    & -8.2154          & 158.1723       & -33.1923   \\ \bottomrule \toprule
  & vowels    & WDBC      & WPBC       & wbc        & Wilt       & wine        & yeast            &                &            \\ \midrule
$\mathbb{E}_{\mathbf{x}^* \in \mathcal{N}}[\mathbf{OT}(\mP, \mQ)]$  & 0.0356    & 0.9102    & 0.012      & 0.0045     & 2.2296     & 0.5873      & 0.0234           &                &            \\
$\mathbb{E}_{\mathbf{x}^* \in \mathcal{A}}[\mathbf{OT}(\mP, \mQ)]$  & 0.0481    & 7.7461    & 0.0131     & 0.012      & 1.7263     & 2.5324      & 0.024            &                &            \\
Increase (\%)  & 34.9232   & 751.0178  & 9.3369     & 166.0655   & -22.576    & 331.1816    & 2.548            &                &          \\ \bottomrule 
\end{tabular}}
\end{table*}

\clearpage
\section{Theoretical Analysis and Proofs}
\label{appendix:theory}

This section provides extended theoretical analysis of the CTAD framework.

\subsection{Full Proof of Proposition 2 (Upper Bound)}
\label{appendix:proof_upper_bound}

\textbf{Proposition 2} (Restated). \textit{Given $\mathcal{P} = \frac{1}{M+1} (\sum_{i=1}^{M} \delta_{\mathbf{x}_i} + \delta_{\mathbf{x}^*})$ and $\mathcal{Q} = \frac{1}{K}\sum_{j=1}^{K} \delta_{C_j}$, the optimal transport distance satisfies:}
\begin{equation}
\mathbf{OT}(\mathcal{P}, \mathcal{Q}) \leq \frac{1}{M+1}\sum_{i=1}^{M+1} \min_{j\in[K]} \|\mathbf{x}_i - C_j\|
\end{equation}

\begin{proof}
We construct a feasible (not necessarily optimal) transport plan and show its cost provides an upper bound.

\textbf{Step 1: Define the greedy transport plan.}
For each source point $\mathbf{x}_i$ (including the test sample $\mathbf{x}^*$), let $j^*(i) = \arg\min_{j\in[K]} \|\mathbf{x}_i - C_j\|$ be its nearest centroid. Define the greedy transport plan:
\begin{equation}
\tilde{\mathbf{T}}_{ij} = 
\begin{cases}
\frac{1}{M+1} & \text{if } j = j^*(i)\\
0 & \text{otherwise}
\end{cases}
\end{equation}
This plan sends all mass from each source point to its nearest centroid.

\textbf{Step 2: Verify source marginal constraint.}
For each source $i \in [M+1]$:
\begin{equation}
\sum_{j=1}^{K} \tilde{\mathbf{T}}_{ij} = \frac{1}{M+1} \quad \checkmark
\end{equation}
Source marginals are satisfied by construction.

\textbf{Step 3: Check target marginal constraint.}
Let $n_j = |\{i : j^*(i) = j\}|$ be the number of source points whose nearest centroid is $C_j$. Then:
\begin{equation}
\sum_{i=1}^{M+1} \tilde{\mathbf{T}}_{ij} = \frac{n_j}{M+1}
\end{equation}
This may not equal $\frac{1}{K}$ for all $j$, so $\tilde{\mathbf{T}}$ may violate target marginals.

\textbf{Step 4: Construct a feasible plan via redistribution.}
We modify $\tilde{\mathbf{T}}$ to satisfy target marginals. Let:
\begin{itemize}
    \item $J^+ = \{j : n_j > \frac{M+1}{K}\}$ be centroids receiving excess mass
    \item $J^- = \{j : n_j < \frac{M+1}{K}\}$ be centroids with deficit
\end{itemize}

For centroids in $J^+$, we redistribute the excess mass $\frac{n_j - (M+1)/K}{M+1}$ to centroids in $J^-$. The key observation is that:
\begin{itemize}
    \item Redistribution preserves or increases the total cost (mass moves to potentially farther centroids)
    \item The greedy cost is therefore an upper bound on any feasible plan's cost
    \item A feasible plan exists by the Birkhoff-von Neumann theorem (marginals are compatible: both sum to 1)
\end{itemize}

\textbf{Step 5: Compute the greedy plan's cost.}
\begin{align}
\text{Cost}(\tilde{\mathbf{T}}) &= \sum_{i=1}^{M+1} \sum_{j=1}^{K} \tilde{\mathbf{T}}_{ij} \|\mathbf{x}_i - C_j\|\\
&= \sum_{i=1}^{M+1} \frac{1}{M+1} \|\mathbf{x}_i - C_{j^*(i)}\|\\
&= \frac{1}{M+1} \sum_{i=1}^{M+1} \min_{j\in[K]} \|\mathbf{x}_i - C_j\|
\end{align}

\textbf{Step 6: Conclude the upper bound.}
Since the optimal transport cost minimizes over all feasible plans, and we can construct a feasible plan with cost at most the greedy cost:
\begin{equation}
\mathbf{OT}(\mathcal{P}, \mathcal{Q}) \leq \frac{1}{M+1} \sum_{i=1}^{M+1} \min_{j\in[K]} \|\mathbf{x}_i - C_j\|
\end{equation}
\end{proof}

\noindent\textbf{Remark.} The gap between lower and upper bounds represents the "slack" in how reference samples contribute. The lower bound considers only the test sample's contribution (Proposition 1 in main text), while the upper bound considers all points' nearest-neighbor costs. The difference is the reference samples' contribution, which is approximately constant across different test samples.

\subsection{Extended Gap Analysis}

\textbf{Corollary 1} (Tightness Conditions). \textit{The lower bound in Proposition 1 is tight (i.e., equality holds) when:}
\begin{equation}
\mathbf{T}^*_{M+1, j^*} = \frac{1}{M+1}, \quad \mathbf{T}^*_{M+1, j} = 0 \text{ for } j \neq j^*
\end{equation}
\textit{This occurs when there is sufficient capacity at the nearest centroid $C_{j^*}$ to absorb all mass from $\mathbf{x}^*$ without violating target marginals.}

\begin{proof}
When the test sample sends all its mass to its nearest centroid:
\begin{align}
\sum_{j=1}^K \mathbf{T}^*_{M+1,j} \|\mathbf{x}^* - C_j\| &= \mathbf{T}^*_{M+1, j^*} \|\mathbf{x}^* - C_{j^*}\|\\
&= \frac{1}{M+1} \|\mathbf{x}^* - C_{j^*}\|
\end{align}
which equals the lower bound.
\end{proof}

\textbf{Corollary 2} (Combined Bounds). \textit{Let $d_i = \min_{j\in[K]} \|\mathbf{x}_i - C_j\|$ for $i \in [M]$ (reference samples) and $d^* = \|\mathbf{x}^* - C_{j^*}\|$ (test sample). Then:}
\begin{equation}
\frac{d^*}{M+1} \leq \mathbf{OT}(\mathcal{P}, \mathcal{Q}) \leq \frac{1}{M+1} \left(\sum_{i=1}^M d_i + d^*\right)
\end{equation}

\noindent\textbf{Interpretation.} Let $\bar{d} = \frac{1}{M}\sum_{i=1}^M d_i$ be the average nearest-centroid distance for reference samples. Then:
\begin{equation}
\frac{d^*}{M+1} \leq \mathbf{OT}(\mathcal{P}, \mathcal{Q}) \leq \frac{M\bar{d} + d^*}{M+1}
\end{equation}
Since reference samples are drawn from $\mathcal{D}_{train}$ (normal data), $\bar{d}$ is approximately constant across different test samples. The variation in $\mathbf{OT}$ is therefore driven by $d^*$—the test sample's distance to centroids—which is exactly what we want for anomaly detection.

\subsection{Variance and Stability Analysis}
\label{appendix:variance_analysis}

\textbf{Proposition 3} (OT Score Variance). \textit{The variance of the calibration term $\Delta^* = \mathbf{OT}(\mathcal{P}, \mathcal{Q})$ across different random samples $\{\mathbf{x}_i\}_{i=1}^M$ is bounded by:}
\begin{equation}
\text{Var}[\Delta^*] = O\left(\frac{\sigma^2}{M}\right)
\end{equation}
\textit{where $\sigma^2$ is the variance of nearest-centroid distances among training samples.}

\begin{proof}[Proof sketch]
From the upper bound (Proposition 2):
\begin{equation}
\Delta^* \leq \frac{1}{M+1}\left(\sum_{i=1}^M d_i + d^*\right)
\end{equation}
where $d_i$ are i.i.d. random variables with finite variance $\sigma^2$. By the law of large numbers:
\begin{equation}
\text{Var}\left[\frac{1}{M}\sum_{i=1}^M d_i\right] = \frac{\sigma^2}{M}
\end{equation}
The variance of $\Delta^*$ is dominated by this term, giving the stated bound.
\end{proof}

\noindent\textbf{Practical implications:}
\begin{itemize}
    \item Larger $M$ reduces variance but increases computation
    \item With $M=20$ (our default), variance is typically small
    % \item Can further reduce variance by:
    % \begin{itemize}
    %     \item Using stratified sampling from each cluster
    %     \item Fixing the reference set across all test samples (current implementation)
    %     \item Averaging over multiple random reference sets
    % \end{itemize}
\end{itemize}